\pdfoutput=1
\documentclass{article}

     \PassOptionsToPackage{square, numbers, compress}{natbib}

     \usepackage[preprint]{neurips_2020}

\usepackage[utf8]{inputenc} 
\usepackage[T1]{fontenc}    
\usepackage{hyperref}       
\usepackage{url}            
\usepackage{booktabs}       
\usepackage{amsmath,amsfonts}       
\usepackage{nicefrac}       
\usepackage{microtype}      
\usepackage{algpseudocode,algorithm,algorithmicx}
\usepackage{graphicx}
\usepackage{subcaption}
\usepackage{comment}
\usepackage{caption}
\usepackage{cleveref,xcolor}

\crefformat{section}{\S#2#1#3} 
\crefformat{subsection}{\S#2#1#3}
\crefformat{subsubsection}{\S#2#1#3}

\newcommand{\Or}{\mathcal{O}}
\newcommand{\brand}{A2SGD }
\newcommand{\Expect}[1]{{I\kern-.3em E \{#1\}}}
\newcommand{\Alpha}{{\mathrm{\mathbb{A}}}}
\newcommand{\Beta}{{\mathrm{\mathbb{B}}}}
\newcommand{\Prob}{{\mathrm{\mathbb{P}}}}
\newcommand{\Aop}{{\mathit{\mathbf{enc}}}}
\newcommand{\specialcell}[2][c]{\begin{tabular}[#1]{@{}c@{}}#2\end{tabular}}
	
\begin{document}
\title{$\Or(1)$ Communication for Distributed SGD through Two-Level Gradient 
Averaging}

%

\author{%
  Subhadeep Bhattacharya \\
  Florida State University\\
  \texttt{bhattach@cs.fsu.edu} \\
  \And
  Weikuan Yu \\
  Florida State University \\
  \texttt{yuw@cs.fsu.edu} \\
  \AND
  Fahim Tahmid Chowdhury \\
  Florida State University \\
  \texttt{fchowdhu@cs.fsu.edu} \\
}

\maketitle
\begin{abstract}
Large neural network models present a hefty communication challenge to distributed 
Stochastic Gradient Descent (SGD), with a communication complexity 
of $\Or(n)$ per worker for a model of $n$ parameters. 
Many sparsification and quantization techniques have been proposed
to compress the gradients, 
some reducing the communication complexity to $\Or(k)$, where $k \ll 
n$. 
In this paper, we introduce a strategy called two-level gradient averaging (A2SGD) 
to consolidate all gradients down to merely two local averages per worker before
the computation of two global averages for an updated model. 
A2SGD also retains local errors to maintain the variance for fast convergence.
Our theoretical analysis shows that A2SGD converges similarly like the default distributed SGD algorithm. 
Our evaluation validates the theoretical conclusion and 
demonstrates that A2SGD significantly reduces the communication traffic
per worker, and improves the overall training time of LSTM-PTB by $3.2\times$ and 
$23.2\times$, respectively, compared to Top-K and QSGD. 
To the best of our knowledge, A2SGD is the first to achieve $\Or(1)$
communication complexity per worker for distributed SGD. 
   
\end{abstract}

\section{Introduction}
\label{sec:intro}

Deep learning has found great success in image
classification, speech recognition, and language 
processing~\cite{szegedy2015going, resnet1k}, etc.  The demand for more 
powerful
and accurate Deep Neural Networks (DNNs) leads to large and complex models
with more than 1 Billion parameters, such as GPT-2 (1.5B)~\cite{GPT2} and
Transformer (6B)~\cite{Transformer6B}.  Such large-scale models require
distributed Stochastic Gradient Descent (SGD) algorithms for training.
Distributed SGD typically adopts data parallelism, in which 
$P$ workers hold the same model $w \in \mathbb{R}^n$ of $n$ parameters and train it in parallel
through many iterations. 
At the $t$-th iteration, weight $w$ is updated as follows based on the learning 
rate $\eta$ and the gradients $g$:
\vspace{-0.5pc}
\begin{equation}
w_{t+1} = w_t - \eta_t \frac{1}{P} \sum_{p=1}^{P}g^p_t,
\label{equ:dsgd}
\end{equation}
where a worker computes local 
gradients (of the same size $n$) for the model using its fraction of a mini-batch,
and exchanges the gradients across all workers for an updated global model.
Such a global exchange and synchronization problem imposes a hefty requirement on both the latency and
bandwidth of distributed systems, and hampers the scalability of distributed 
SGD~\cite{agarwal2011distributed, 
strom2015scalable, NIPS13SML, OSDI14SDML, OSDI14Adam}.
Various strategies have been proposed to tackle this problem by increasing the mini-batch 
sizes~\cite{goyal2017accurate, you2017scaling, li2014efficient},
reducing the rounds of 
communication~\cite{baiduallreduce,awan:ccgrid19,jiang2018linear}, 
or pruning the neural 
networks~\cite{NIPS1989_OBD,NIPS1992_OBS,NIPS2016_DNS,ICLR2016DC}. 

Particularly, there exists a fundamental bottleneck, i.e., the need to 
transfer $\Or(n)$ local gradients for each worker.
Many studies have proposed to compress the gradients through 
quantization~\cite{wen2017terngrad,alistarh2017qsgd,karimireddy2019error,
bernstein2018signsgd,strom2015scalable,jiang2018linear} and/or
sparsification~\cite{alistarh2018convergence,shi2019_GaussianTopK,aji-heafield-2017-sparse,stich2018sparsified}.
Quantization enables lossy compression of gradients by 
reducing the precision of their representation to a varying degree, 
from 1BitSGD~\cite{seide20141,strom2015scalable} with only a sign bit, 
TernGrad~\cite{wen2017terngrad} with three numerical levels \{-1, 0, 1\},
to QSGD~\cite{alistarh2017qsgd} that supports multiple quantization levels.
These quantization techniques can reduce the communication volume by at most
32 times, assuming gradients are single-precision float-point numbers. 

Gradient sparsification can achieve higher compression 
by selecting only $k$ out of $n$ gradients to reduce the communication traffic per worker
~\cite{shi2019_GaussianTopK,aji-heafield-2017-sparse,stich2018sparsified}.
The selection criteria can be based on a user-defined threshold 
(Top-K)~\cite{aji-heafield-2017-sparse}, 
a gaussian-estimated threshold (Gaussian-K) ~\cite{shi2019_GaussianTopK},
or simple randomization (Rand-K)~\cite{stich2018sparsified}. 
Prior results~\cite{stich2018sparsified,alistarh2018convergence}
have shown that, theoretically, sparsified SGD can converge within the same upper bound
as the original distributed SGD (dense SGD) algorithm, which exchanges full gradients.
In practice, they have different convergence behaviors, for which 
Shi et~al.~\cite{shi2019_GaussianTopK} have performed a theoretical analysis 
to distinguish them. 

In this paper, we propose a novel algorithm 
different from both sparsification and quantization. Our algorithm
\textit{two-level gradient averaging (A2SGD)} 
consolidates all local gradients down to merely two local means and 
achieves \textit{a communication complexity of $\Or(1)$ per worker}.
It then aggregates the local means into two global means
across all workers for an updated model.
The key idea behind A2SGD is to not drop or quantize any  
gradient, but average all 
local gradients and record the difference between the gradients and
the resulting means locally at each worker. 
In doing so, A2SGD retains local errors to maintain the
same variance across gradients as dense SGD, avoiding 
any potential variance blowup or any increase on the number of iterations.
A2SGD does not require complex sampling or sorting of gradients but
only simple calculations for the two means and their differences with the gradients.
Our theoretical analysis shows that A2SGD converges similarly like dense SGD. 
Our evaluation validates the theoretical conclusion and demonstrates that 
A2SGD significantly improves
the execution time per iteration and the overall training time, 
by $3.2\times$ and $23.2\times$, respectively, compared to Top-K and QSGD 
for LSTM-PTB, a big model with around 66 million parameters. 
Compared to the default dense SGD algorithm, 
A2SGD improves the overall training time of LSTM-PTB by $1.72\times$. 
Compared to other techniques such as Top-K, Gaussian-K and QSGD, 
A2SGD achieves the best overall performance in terms of convergence
accuracy, execution time, and scaling efficiency.

\noindent
\textbf{Our Contributions.}
In summary, by examining the scalability challenge of gradient
synchronization in distributed SGD and analyzing its computation and communication complexities,
we have proposed a two-level gradient averaging algorithm A2SGD for distributed workers 
to exchange only two means globally.
Our theoretical analysis and experimental results have confirmed the convergence of A2SGD 
and demonstrated that A2SGD achieves an overall improvement compared to 
other sparsification and quantization algorithms~\cite{shi2019_GaussianTopK,stich2018sparsified,alistarh2017qsgd}. 
Our results also show that A2SGD achieves fast computation complexity. 
To the best of our knowledge, A2SGD is the first to achieve $\Or(1)$
communication complexity per worker for distributed SGD. 
It can be further integrated with 
quantization~\cite{seide20141,strom2015scalable,wen2017terngrad,alistarh2017qsgd,wu2018error,
 karimireddy2019error,tang2019doublesqueeze,haddadpour2019trading} 
or communication reduction~\cite{jiang2018linear,baiduallreduce}
techniques for additional improvements on the scalability of distributed SGD.

\section{Related Work}

\noindent
\textbf{Gradient Quantization.} Gradient quantization takes
advantage of the fact that 
distributed SGD can still converge with low-precision 
gradients instead of 32-bit floating-point representations.
A wide variety of quantization techniques have tried to represent gradients
in 16 bits~\cite{NIPS2017_flexpoint,narang2018mixed,jia2018highly}, 8
bits~\cite{NIPS2018_wang8bit}, 2.8 bits~\cite{alistarh2017qsgd}, 2 bits~\cite{Choi2019ACCURATEAE}, 
or even 1 bit~\cite{seide20141,strom2015scalable}.
In addition, Wen et al.~\cite{wen2017terngrad} have quantized gradients from workers to the server using ternary values \{-1, 0, 1\}.
Furthermore, some studies have provided theoretical analysis on the convergence 
guarantees 
of quantization techniques~\cite{alistarh2017qsgd,wu2018error,jiang2018linear,
karimireddy2019error,tang2019doublesqueeze,haddadpour2019trading}. 
Notwithstanding the compulsory cost for quantizing the gradients, 
quantization is inherently limited by its optimization scope, i.e., 
the number of bits representing the gradients.
Thus it can reduce the network traffic by at most 32x compared to 32-bit numbers. 
The overall improvement on the time per iteration or the total training time
is further limited. 

\noindent
\textbf{Gradient Sparsification.} 
Compared to gradient quantization, 
sparsification examines the total number ($n$) of gradients
and selectively transfers only a small number ($k$) of them 
while still allowing DNN models to converge.  Because $k$ can be several orders 
of magnitude smaller than $n$, sparsification 
techniques~\cite{strom2015scalable,aji-heafield-2017-sparse,
lin2017deep,wangni2018gradient,stich2018sparsified,alistarh2018convergence,
jiang2018linear,shi2019convergence} have been shown to be much more effective 
than quantization in reducing the communication traffic.
Several studies~\cite{strom2015scalable,aji-heafield-2017-sparse}
have differentiated gradient values by magnitude and purged the small ones
under a threshold. \citet{lin2017deep} adopted a number of optimizations 
to achieve very high sparsity in the exchanged gradients and carefully 
tuned the hyperparameters of DNN models to avoid any loss of accuracy.
\citet{wangni2018gradient,stich2018sparsified,alistarh2018convergence}
theoretically analyzed the performance of sparsification 
and established various bounds on the convergence rate. 
Nonetheless, it is imperative for these techniques to process all gradients,
at certain computation costs, to reach their desired sparsity levels. 
As an alternative,~\citet{shi2019_GaussianTopK} 
have recently proposed to take advantage of the gaussian 
distribution property of gradients and statistically pinpoint a threshold 
to select the top $k$ gradients at low computation costs.

\section{Design of Distributed SGD with Two-Level Gradient Averaging}
\label{subsec:gradsync}

As mentioned in~\cref{sec:intro}, gradient synchronization imposes a
fundamental scalability challenge for data-parallel distributed SGD due to the
requirement for all workers to exchange their gradients.
While the sparsity and quantization levels are important to
the communication complexity of gradient synchronization, 
the computation efficiency of sparsification and quantization can be critical
to the scalability of gradient synchronization as well.
\citet{shi2019_GaussianTopK} reported that, while Top-K sparsification reduces the communication traffic, 
its computation overhead can offset the overall benefit, resulting in a
suboptimal improvement on the execution time per iteration.
On systems with high-bandwidth communication networks at 100 Gbps or higher, 
the computation costs from Top-K sparsification can overshadow its gains on
communication efficiency, as we have observed in our experimental
evaluation (\cref{sec:eval}). The same tradeoff happens to quantization techniques such as QSGD.
\citet{shi2019_GaussianTopK} proposed Gaussian-K
to avoid costly sorting and selection of top K elements across all gradient values.
Gaussian-K assumes a gaussian distribution of gradient values and 
estimates a statistical threshold for the selection of gradient values.
It has demonstrated the importance of low computation for sparsification.
Sparsification and quantization can also be combined and generalized
as compression techniques for the improvement of gradient
synchronization~\cite{jiang2018linear,SC19_SparCML,karimireddy2019error}.
All these studies have mitigated the computation costs of gradient
costs while allowing the models to converge. But all of them 
require the workers in distributed SGD to exchange some fraction of their gradients. 

\begin{figure}
\vspace{-1pc}
\begin{minipage}{.61\textwidth}
  \centering
  \begin{subfigure}[b]{0.495\textwidth}
          \centering
          \includegraphics[width=\textwidth]{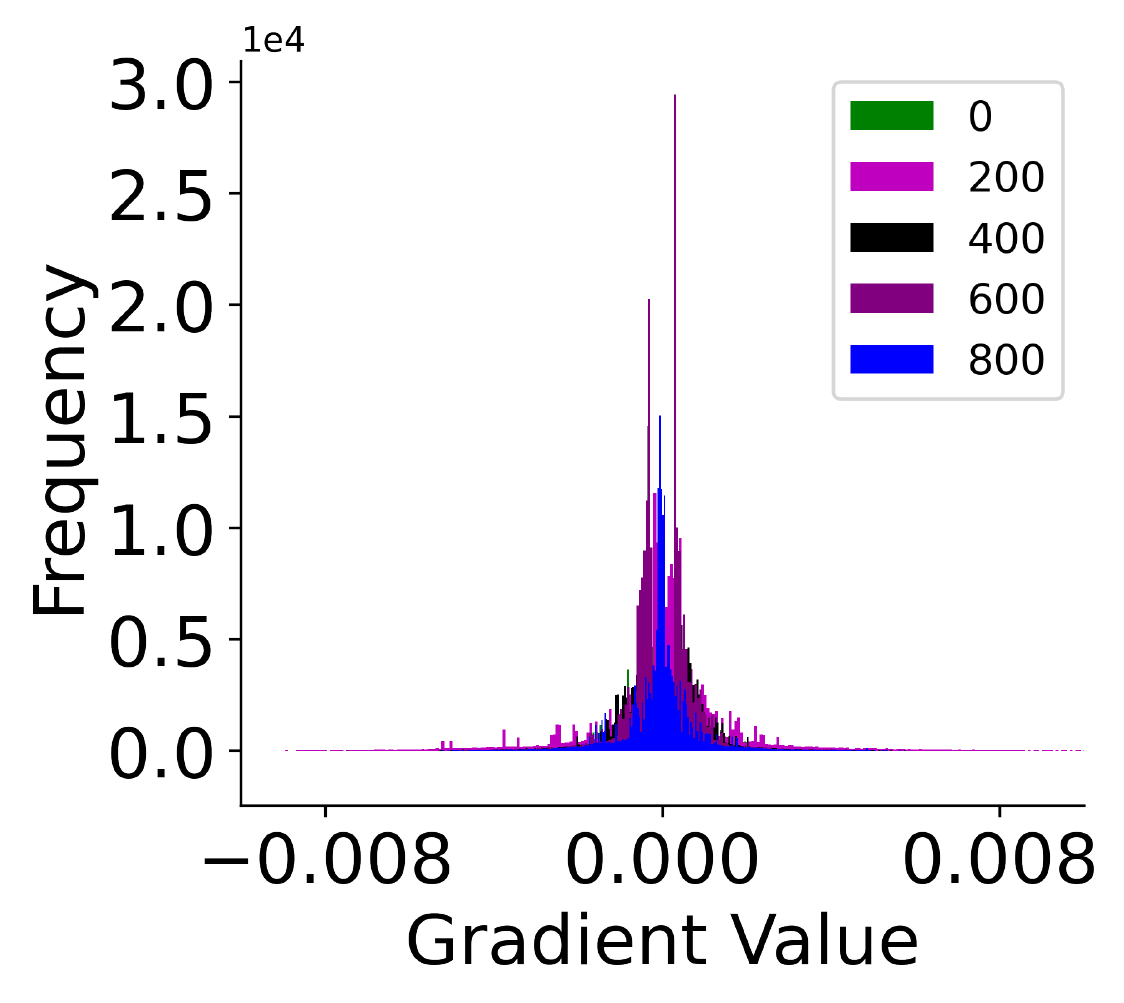}
          \caption{FNN3}
  \end{subfigure}%
  \hfill%
  \begin{subfigure}[b]{0.495\textwidth}
          \centering
          \includegraphics[width=\textwidth]{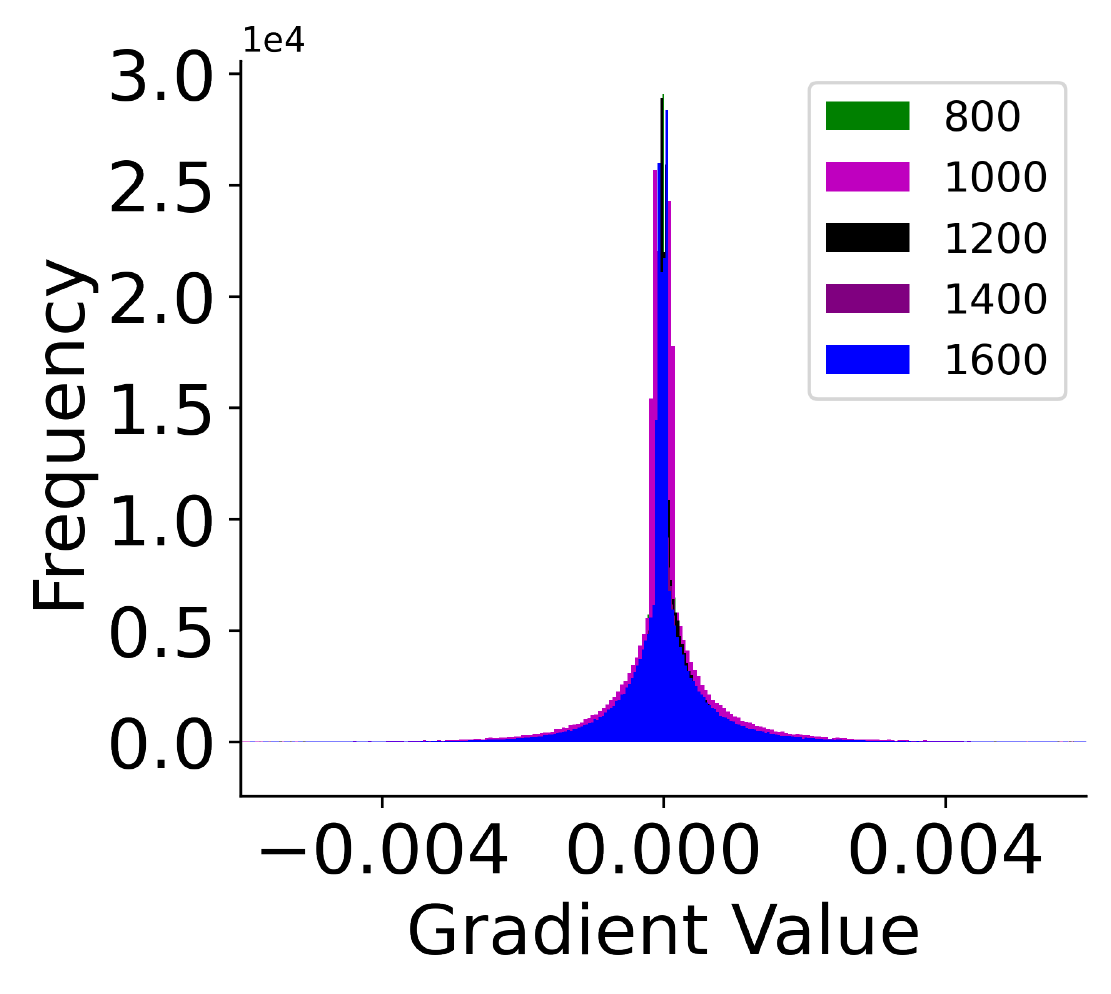}
          \caption{ResNet20}
  \end{subfigure}
  \caption{Progression of Gradient Distribution.}
  \label{fig:hist_density}
\end{minipage}
\begin{minipage}{.39\textwidth}
     \centering
     \includegraphics[width=\textwidth]{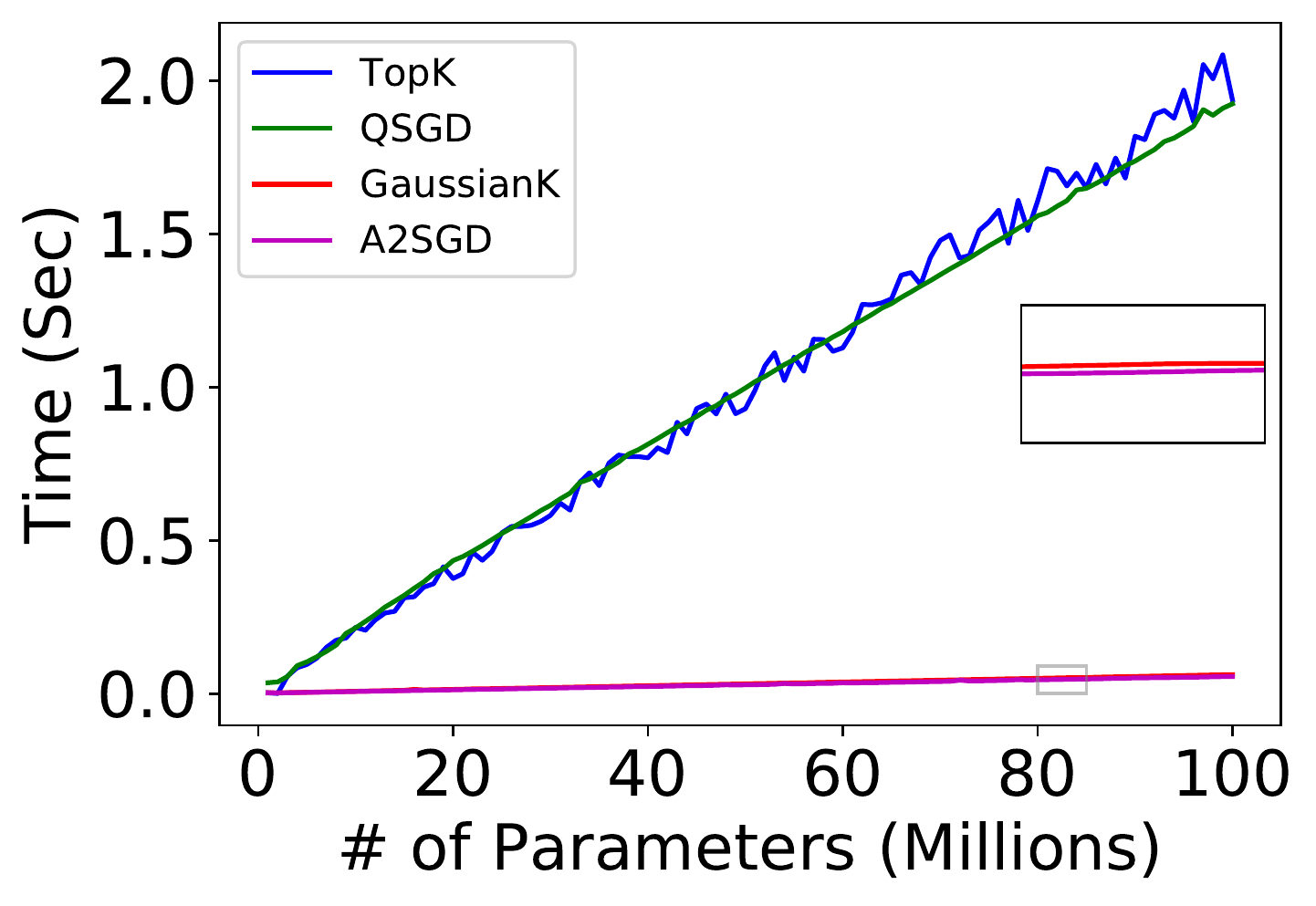}
     \caption{Comparison of A2SGD computation time with other algorithms.}
    \label{fig:computation_time}
\end{minipage}
\vspace{-1.0pc}
\end{figure}

We propose an alternative to sparsification. Instead of 
selecting a top fraction of gradient values, we can exchange the mean 
across the distributed workers.
To avoid over simplification caused by a unified mean, we arrange the gradient
values into two groups: positive ($\ge 0$) and negative, and compute their absolute 
means accordingly. Then all workers can exchange these two means for
gradient synchronization. 
A global negative mean is then computed by averaging the negative means from
all workers; and a global positive mean by averaging the positive means.
We refer to our algorithm as~\textit{Two-level Gradient Averaging (A2SGD)}. 
It effectively reduces the communication traffic 
down to two values, achieving the communication complexity of $\Or(1)$.
While this is different from gradient clipping 
in prior studies~\cite{wen2017terngrad,lin2017deep,karimireddy2019error},
it would also lead to some distorted gradients for the workers.
Several prior studies~\cite{karimireddy2019error, seide20141, strom2015scalable,
stich2018sparsified,wu2018error} pursued the idea of error-feedback to
correct either the momentum or the variance, or both to improve the
convergence accuracy of training. 
To limit the impact on variance, we equip A2SGD with a similar error-feedback mechanism through a local error
vector. We compute the difference between the gradients and the two means, and use
a local error vector to store the difference. 
When the global synchronization completes, the error vector is added back to the global means,
according to the corresponding positions of the original positive and negative gradients,
to generate the updated gradients.

Figure~\ref{fig:hist_density} shows the frequency distribution of gradients
and its progression with an increasing number of iterations, 
for two representative models: FNN-3 and ResNet-20. 
Most of the values are close on either side of zero, following a normal distribution. 
Besides, as the models finish more iterations of the training, 
more gradient values are converging to the center around zero. 
These distribution plots from one representative worker provide a pictorial 
visualization 
on the convergence of gradients across all workers. 

For an initial assessment on A2SGD's computation cost, we have measured its computation time 
with an increasing number of parameters, compared to Top-K, Gaussian-K and QSGD. 
A2SGD, Top-K, Gaussian-K use PyTorch~\cite{PyTorch} APIs with GPU support, 
   and QSGD is implemented in Numpy~\cite{git_qsgd}
(see~\cref{subsec:setup} for more details on our experimental setup).
Figure~\ref{fig:computation_time} shows that A2SGD and Gaussian-K have much
lower computation complexity than QSGD and Top-K. 
A2SGD has a slightly lower
computation cost than Gaussian-K because Gaussian-K has to estimate a
threshold~\cite{shi2019_GaussianTopK} before gradient selection. We have 
elaborated the detailed computation complexity for these algorithms 
in~\cref{subsec:complexity}. 
These initial results on the computation time suggest that A2SGD is very promising to support efficient
gradient synchronization because of its significant reduction on communication traffic 
at very low computation costs. 

\subsection{Details of A2SGD}

For a gradient vector $v = \{v_1, v_2, ..., v_n\} \in \mathbb{R}^n$, 
we denote $\mu_{+} (v)  = \Expect{(v_i)} \; \forall \; v_i \ge 0 $ as  the
absolute mean of all positive values $v_i$ in $v$, 
and $\mu_{-} (v) = \Expect{|v_i|} \; \forall \; v_i < 0 $ the absolute mean of all negative values in $v$.  
We introduce a new operator $\Aop$ below to transform the values of $v$.
\vspace{-0.5pc}
\begin{equation}
\Aop (v) = pos(v) \cdot \mu_{+} (v) - neg(v) \cdot \mu_{-} (v) 
\end{equation}
where $pos(v)$ and $neg(v) \in \mathbb{R}^n$ are vectors with values $\in \{1, 0\}$. 
The former has 1 in the corresponding positions $\forall \; v_i \ge 0, i \in [1,n]$, 
and the latter with 1, $\forall \; v_i < 0, i \in [1,n]$,

\begin{algorithm}
	\caption{Parallel A2SGD Algorithm at Worker p}
	\label{alg:A2SGD}
	\textbf{Input:} Initial learning rate $\eta_0$, weight $w_0 = \vec{0}$ 
	\begin{algorithmic}[1]
		\For{$t \gets 1$ to $T-1$}
		\State $g_t \gets SGD_t^p$ through training with $M^p_t$
		\algorithmiccomment{training $M^p_t$, portion of mini-batch for p.}
		\label{opt01}
		\State $\mu_{t,+} \gets \mu_+(g_t)$ and $\mu_{t,-} \gets \mu_-(g_t)$ \label{opt02}
		\State $\epsilon_t \gets g_t - \Aop(g_t)$ 
		\algorithmiccomment{Store the errors in a local vector $\epsilon_t$}
                \label{opt03}
		\State $(\bar{\mu}_{t,+}, \bar{\mu}_{t,-}) \gets
		\textbf{Allreduce}((\mu_{t,+}, \mu_{t,-}),\textbf{average})$ 
		\algorithmiccomment{Global exchange of two means}
                \label{opt04}
		\State $g_t \gets \epsilon_t +
		pos(g_t) \cdot \bar{\mu}_{t,+} -
		neg(g_t) \cdot \bar{\mu}_{t,-} $ 
		\algorithmiccomment{Aggregate the error vector $\epsilon_t$.}
		\label{opt05}
		\State $w_{t+1} \gets w_{t} -\eta_t \cdot g_t$ 
		\algorithmiccomment{Update the model.}
		\label{opt06}
		\EndFor
		\State $g_{T-1} \gets
		\textbf{Allreduce}(g_{T-1}, 
		\textbf{average})$ \label{opt07}
		\State $w_{T} \gets w_{T-1} -\eta_{T-1} \cdot g_{T-1}$ 
		\label{opt08}
	\end{algorithmic}
\end{algorithm}
\vspace{-0.5pc}

Algorithm~\ref{alg:A2SGD} describes the proposed A2SGD algorithm in detail. 
Worker p starts with a learning rate $\eta_0$ and an initial weight $w_0$.
At any iteration $t$, Worker p computes its stochastic gradients $g_t$ 
by training $SGD^p_t$ 
with its portion of mini-batch $M^p_t$ (Line~\ref{opt01}).
It then extracts the means for positive and negative gradients 
(Line~\ref{opt02}) and subtracts the vector constructed from the means (Line~\ref{opt03}). 
The errors are stored in a local error vector $\epsilon_t$.
All workers call the Allreduce operation to exchange their local means
and get back the global means (Line~\ref{opt04}). 
The global means are then combined with the errors stored in~$\epsilon_t$
into the new gradients $g_t$ (Line~\ref{opt05}).
Finally, the model weight is updated at Line~\ref{opt06} using the new gradient and
the current learning rate.  
At the end of the training iterations, one more iteration is performed 
to synchronize the model across all workers (Lines~\ref{opt07}
and~\ref{opt08}).

\subsection{Convergence Analysis}

Distributed SGD can be analyzed in the framework of online learning systems. 
The convergence analysis of \brand follows the convergence proof of GOGA in 
\citet{bottou1998online}, similarly as previous studies~\cite{wen2017terngrad}. 
Assumption 1, Assumption 2 and Lemma 1 are also adapted 
from the same.

\noindent
\textbf{Assumption 1.} $C(w)$ has a single minimum $w^*$ and gradient 
$-\nabla (w)$ always points to $w^*$
\begin{equation}
\forall\epsilon > 0, \inf_{\parallel w-w^*\parallel > \epsilon}
(w-w^*)^T\nabla_wC(w) > 0\label{eq:5}
\end{equation}
Convexity is a subset of Assumption 1, and we can easily find \textit{{non-convex}} functions satisfying it.

\noindent
\textbf{Assumption 2.} Learning rate $\eta_t$ is positive and constrained as follows:
\begin{equation}
\begin{cases}
& \sum_{t=0}^{+\infty}{\eta_t}^2 \; < \; +\infty\\
& \sum_{t=0}^{+\infty}\eta_t \; = \; +\infty\\
\end{cases} \label{eq:6}
\end{equation}
These constraints ensure that $\eta_t$ changes neither very fast nor very slow. 
We define the square of distance between the current weight $w_t$ and the 
minimum $w^*$ below:
\begin{equation}
h_t \overset{\Delta}{=} {\parallel w_t - w^* \parallel}^2 \label{eq:7}
\end{equation}
where $\parallel . \parallel$ is $l_2$ norm. 
We also define the set of all random variables before step $t$ as follows:
\begin{equation}
\mathcal{D}_t \overset{\Delta}{=} (z_{1 \dots t-1}, b_{1 \dots t-1}) \label{eq:8}
\end{equation}
Under Assumption 1 and 2, using Lyapunov process and Quasi-Martingales 
convergence theorem, ~\citet{bottou1998online} proved the Lemma below.

\noindent
\textbf{Lemma 1.} If $\exists \Alpha, \Beta \; > \; 0 \; s.t.$
\begin{equation}
\Expect{(h_{t+1} - (1 + {\eta_t}^2\Beta)h_t) | \mathcal{D}_t} \le -2{\eta_t}(w_t - 
w^*)^T\nabla_wC(w_t) + {\eta_t}^2\Alpha, \label{eq:9}
\end{equation}
then $C(z,w)$ \textbf{converges almost surely} toward minimum $w^*$
$i.e.\; \Prob(\lim\limits_{t\to+\infty}w_t = w^*) = 1$.

In each iteration, gradient $g_t$ substracts its own
means and then gains back the global means. 
We can denote $\bar{\mu}_{t} = pos(g_t) \cdot \bar{\mu}_{t,+} 
- neg(g_t) \cdot \bar{\mu}_{t,-}$ as the vector composed global means,
and $\nabla \mu_{t} = \bar{\mu}_{t} - \Aop(g_t)$ as the net gain.
We draw another assumption similar to Eq. 4.34 of~\citet{bottou1998online}.

\noindent
\textbf{Assumption 3. (Gradient Bound)}
\begin{equation}
\Expect{\parallel g_t + \nabla \mu_{t} \parallel^2} 
\le \Alpha + \Beta\parallel w-w^* \parallel^2 \label{eq:10}
\end{equation}

\noindent
\textbf{Theorem 1.} When the learning system is updated as follows:
\begin{equation}
w_{t+1} = w_t - {\eta_t}(g_t + \nabla \mu_{t}), \label{eq:13}
\end{equation}
then, it \textbf{converges almost surely} toward minimum $w^*$, $i.e.\; 
\Prob(\lim\limits_{t\to+\infty}w_t = w^*) = 1$.
The proof of Theorem 1 is provided in the Appendix. 

\section{Empirical Results}
\label{sec:eval}

In this section, we first describe our experimental setup and then present
our evaluation results validating the convergence of A2SGD. 
In addition, we compare its performance with
dense SGD (Dense for short), 
two sparsification techniques Top-K~\cite{stich2018sparsified} and Gaussian-K~\cite{shi2019_GaussianTopK}
and one quantization technique QSGD~\cite{alistarh2017qsgd}. Our
performance evaluation covers several aspects, including convergence accuracy,
computation and communication complexities, scaling efficiency, and execution time. 

\subsection{Experimental Setup}
\label{subsec:setup}

\begin{table}[!htb]
\vspace{-1pc}
	\caption{Experimental Setup}
	\label{exp_setup}
	\centering
	\begin{tabular}{llllll}
		\toprule
		Model & Dataset & \# Parameters & BatchSize & LR & Policy\\
		\midrule
		FNN-3 & MNIST & 199,210 & 128 & 0.01 & LS(1 x) + GW + PD\\
		\cmidrule(r){1-6}
		VGG-16 & CIFAR10 & 14,728,266 & 128 & 0.1 & LS(1.5 x) + GW + PD + LARS\\
		\midrule
		ResNet-20 & CIFAR10 & 269,722 & 128 & 0.1 & LS(1 x) + GW + PD\\
		\cmidrule(r){1-6}
		LSTM-PTB & PTB & 66,034,000 & 128 & 22 & PD\\
		\bottomrule
	\end{tabular}
\vspace{-0.5pc}
\end{table}

We have implemented A2SGD on top of PyTorch~\cite{PyTorch} v1.3.0 with 
CUDA~\cite{cuda} v10.1, 
and utilized Horovod~\cite{Horovod} v0.19.1 with 
Allreduce~\cite{thakur2005optimization} 
for data-parallel implementation of different models.  Top-K and
Gaussian-K implementations are adapted from a GitHub repository~\cite{git_gaussianK}.
Both implementations use the PyTorch Tensor API. 
We have adapted the QSGD implementation
from a GitHub implementation~\cite{git_qsgd}.
We have conducted all our experiments with $16$ Nvidia V100 GPUs. Each node in 
this system is equipped with 
256 GB CPU memory and 1 V100 
GPUs per node with 16 GB GPU memory. Furthermore, all nodes in the system are
connected with a high-bandwidth 100-Gbps InfiniBand network. 

In our tests, we have employed four different DNN models,
including (1) FNN-3 which is a Feed-forward Neural Network (FNN) with 
three hidden fully connected layers; 2) two types of Convolutional Neural Networks
(CNNs), i.e., VGG-16 and ResNet-20 using CIFAR10 dataset; and 3) LSTM-PTB, i.e., 
the Long Short Term Memory (LSTM) model using Penn Treebank (PTB) dataset. 
We have used LARS~\cite{you2017scaling}, 
Linear Scaling(LS), Gradual Warmup (GW) and Polynomial Decay (PD) 
of learning rate (LR) for the Large Batch experiments. 
In all figures, we label Top-K and Gaussian-K without the hyphen for brevity.
Table~\ref{exp_setup} lists the detailed hyperparameters for these models.


\subsection{Convergence Accuracy}

To demonstrate the convergence of A2SGD, we run all four models, with 30 epochs 
for FNN-3, 150 epochs for VGG-16 and ResNet-20, and 100 epochs for
LSTM-PTB with a varying number of workers.
We measure the top-1 convergence accuracy for FNN-3, VGG-16 and ResNet-20, and
the perplexity score for LSTM. 



\begin{figure} 
\vspace{-0.5pc}
\centering
	\begin{subfigure}[b]{0.25\textwidth}
		\centering
		\includegraphics[width=\textwidth]{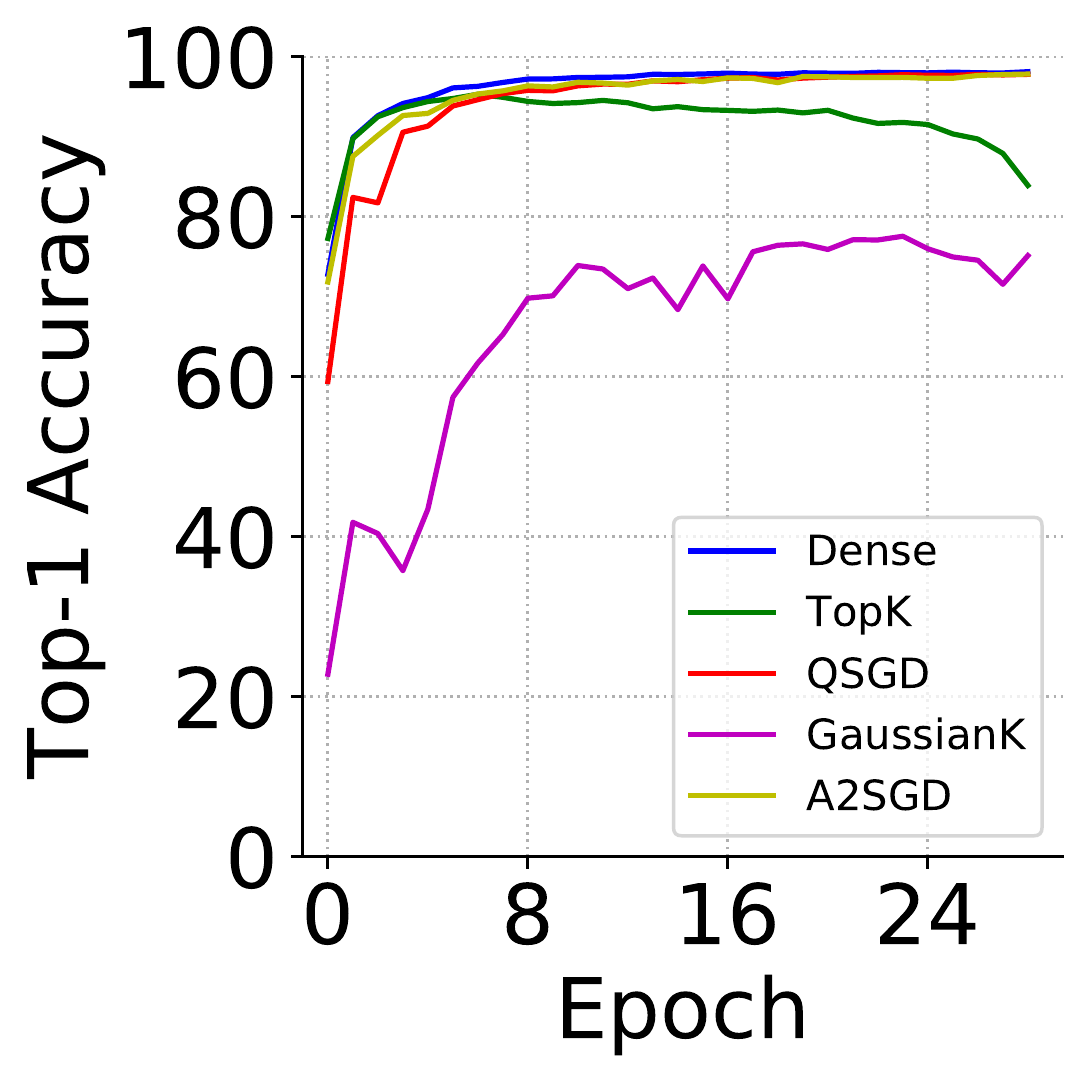}
		\caption{FNN3}
		\label{fig:FNN_acc}
	\end{subfigure}%
        \hspace{-0.5em}
	\begin{subfigure}[b]{0.23\textwidth}
		\centering
		\includegraphics[width=\textwidth]{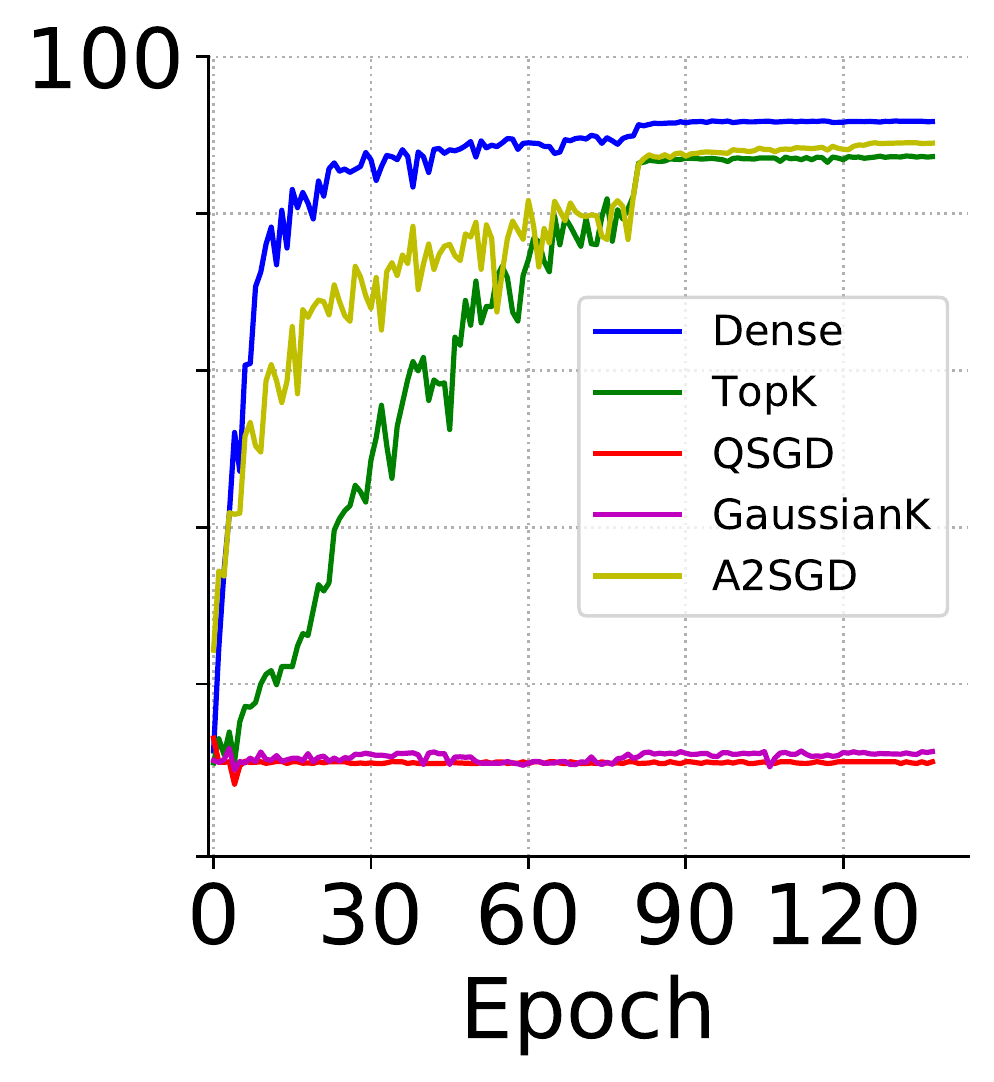}
		\caption{VGG16}
		\label{fig:VGG16_acc}
	\end{subfigure}%
        \hspace{-0.5em}
	\begin{subfigure}[b]{0.24\textwidth}
		\centering
		\includegraphics[width=\textwidth]{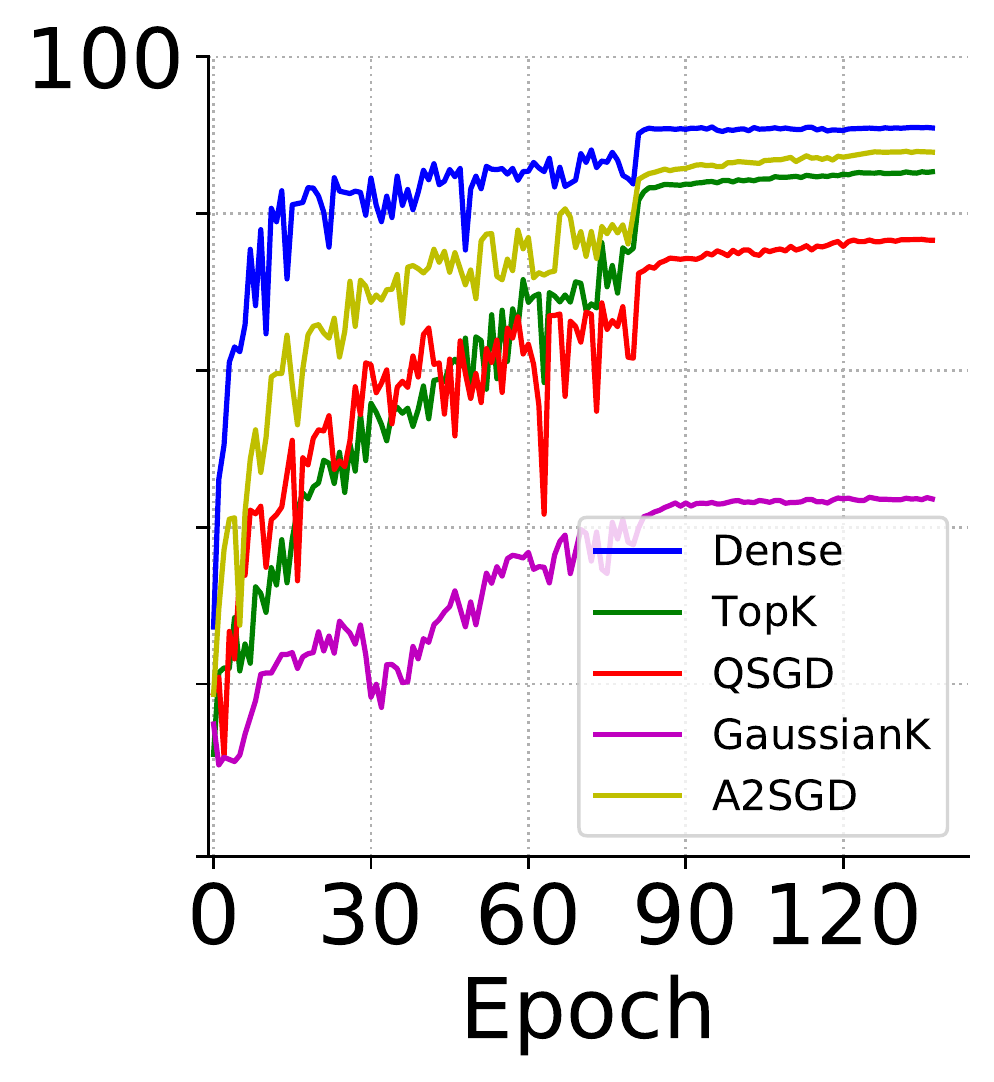}
		\caption{ResNet20}
		\label{fig:ResNet20_acc}
	\end{subfigure}%
        \hspace{-0.5em}
	\begin{subfigure}[b]{0.26\textwidth}
		\centering
		\includegraphics[width=\textwidth]{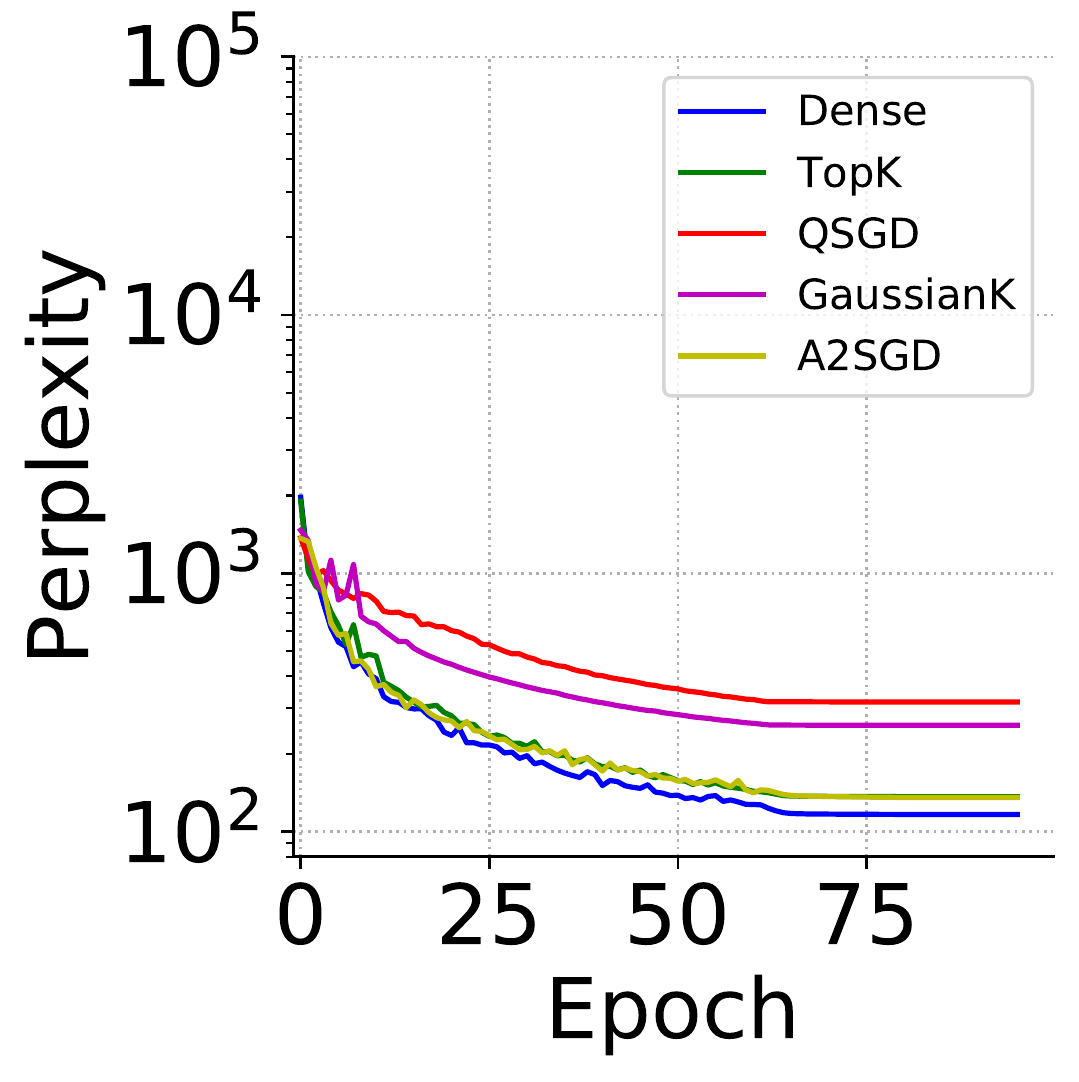}
		\caption{LSTM-PTB}
		\label{fig:LSTM_acc}
	\end{subfigure}%
	\caption{Comparison of Convergence Accuracy with 8 Workers}
\vspace{-1.0pc}
	\label{fig:conv}
\end{figure}

Figure~\ref{fig:conv} shows the convergence performance with 8 workers. The
performance with 2, 4 and 16 workers are included in the Appendix. 
These results show that, for all the cases,~\brand{} achieves the closest
top-1 accuracy to dense SGD within the same number of epochs,
and outperforms the other algorithms in terms of convergence accuracy.
Top-K performs the best overall among the rest of algorithms. 
\brand{} achieves 97.82\%, 87.82\%, 88.80\% top-1 accuracy, 
and 135.53 perplexity for FNN-3, ResNet20, VGG16 and LSTM, respectively. 
In addition, \brand{} achieves 2.5\% and 1.3\% better top-1 accuracies than 
Top-K for ResNet20 and VGG16, respectively. 
Furthermore, Top-K, Gaussian-K, and QSGD all exhibit 
a varying amount of accuracy drops with more workers.

\subsection{Gradient Synchronization Complexities and Scaling Efficiency} 
\label{subsec:complexity}

As discussed in~\cref{subsec:gradsync}, our A2SGD algorithm is designed to improve gradient 
synchronization with reduced communication traffic without costly computation
to process the gradients. To gain an insight on its impact to computation and communication 
in gradient synchronization, we have characterized the asymptotic computation
complexity and the amount of communication traffic (\# bits) per worker 
for A2SGD, in comparison with dense SGD, QSGD, Top-K and Gaussian-K.
In data-parallel distributed SGD, each worker hosts a full copy of the model
and the gradients after each training iteration. 
We assume a model with $n$ parameters, therefore $n$ gradients as well. 

\textbf{Communication Complexity.} In terms of communication traffic, it is evident that dense SGD has to
transfer all gradients from each worker, i.e., $32n$ bits. 
A2SGD transfers two means, i.e., $64$ bits. Top-K and Gaussian-K both 
transfer $k$ gradients, i.e., $32k$ bits.  \citet{alistarh2017qsgd} reported 
that QSGD transfers $2.8n+32$ bits.  Thus A2SGD is the only algorithm that
achieves $\Or(1)$ communication complexity per worker, which can greatly
increase the communication efficiency in the gradient synchronization 
of large DNN models. The amount of communication traffic for these algorithms 
is shown by Column 3 of
Table~\ref{tab:complexity}.

\textbf{Computation Complexity.} 
Dense SGD does not process local gradients, and has a computation complexity of 
$\Or(1)$.
All other algorithms store a local residual or error vector from the
transferred gradients, with a computation complexity of $\Or(n)$.
A2SGD has an overall computation complexity of $\Or(n)$ because, for each 
model,
it traverses all gradients to compute two separate averages, which is still
$\Or(n)$. 
Gaussian-K has a computation complexity of $\Or(n)$ in its formulation 
of the gaussian estimation model as stated in~\cite{shi2019_GaussianTopK}. 
In addition, it has an additional overhead to estimate the threshold based on
its gaussian model.

In the Python implementation without GPU support~\cite{git_qsgd}, QSGD computes 
the second norm (a complexity of $\Or(n)$) and then
applies quantization for each gradient. Thus its total computation complexity is 
$\Or(n^2)$.
A max heap-based implementation of TopK has an overall
computation complexity of $\Or(n + klogn)$, where $n$ is the complexity of 
constructing the max heap and $klogn$ for selecting the largest k elements. 
The computation comparison is shown by Column 2 of Table~\ref{tab:complexity}.
Note that GPU based implementations 
can have higher complexity to facilitate the need of GPU
parallelization~\cite{git_topk,topk_query}.
For both QSGD and Top-K, the cost to
maintain a local error vector is one order lower, and does not change the
overall computation complexity. 

Our analysis of these algorithms in terms of their 
asymptotic computation complexity confirms the superb overall 
efficiency of A2SGD. While achieving $\Or(1)$ communication complexity per 
worker, 
it maintains the minimal asymptotic computation complexity of $\Or(n)$, 
without
the overhead for threshold estimation like Gaussian-K.

\begin{table}[!htb]
\vspace{-0.5pc}
	\caption{Comparison of Gradient Synchronization Complexities and Scaling Efficiency}
	\label{tab:complexity}
	\centering
	\begin{tabular}{lccl}
		\toprule
		Algorithm & \specialcell{Computation\\Complexity} &
                \specialcell{Communication\\(\# bits)} & \specialcell{Scaling 
                Efficiency (8 Workers) \\(FNN/VGG/ResNet/LSTM)}\\
		\midrule
		Dense SGD & $\Or(1)$ & $32n$ 
                & (1.83 / 2.34 / 2.52 / 2.34$\times$) \\ \cmidrule(r){1-4}
		QSGD & $\Or(n^2)$ & $2.8n+32$
                & (1.73 / 0.66 / 2.34 / 0.26$\times$) \\ \cmidrule(r){1-4}
		Top-K & $\Or(n + klogn)$ & $32k$ & (1.76 / 
		2.40 / 1.92 / 1.50$\times$) \\ 
		\cmidrule(r){1-4}
		Gaussian-K & $\Or(n)$ & $32k$ & (1.79 / 2.97 / 2.40 / 6.58$\times$) 
		\\ \cmidrule(r){1-4}
		A2SGD & $\Or(n)$ & $64$ & (1.80 / 3.06 / 2.50  / 6.37$\times$) \\ 
		\bottomrule
	\end{tabular}
\vspace{-0.5pc}
\end{table}

Furthermore, we have evaluated the scaling efficiency of A2SGD in comparison
to the other algorithms.  We measure the throughput of each algorithm 
as the number of images processed per second. Then the scaling efficiency 
is calculated as the normalized throughput with an increasing number of workers. 
Since there is no gradient synchronization for only one worker, we use the
throughput of dense SGD with two workers for normalization in order to
demonstrate synchronization costs.
Specifically, it is calculated as \textit{Scaling Efficiency} = $(t_8 / t_2^D)$, 
where $t_2^D$ is the throughput (average images processed per iteration) of dense SGD with 2
workers, and $t_{8}$ is the overall throughput with 8 workers for any specific algorithm. 
A higher throughput reflects the better efficiency in processing images.
As shown by the last column in Table~\ref{tab:complexity}, A2SGD and
Gaussian-K have better scaling efficiency than the other three algorithms.
The reason Gaussian-K performs comparably to A2SGD is because of its
Allgather implementation of gradient exchange as will be discussed in~\cref{subsec:time}.

\subsection{Execution Time}
\label{subsec:time}

\begin{figure} 
	\centering
\begin{subfigure}[b]{0.25\textwidth}
	\centering
	\includegraphics[width=\linewidth]{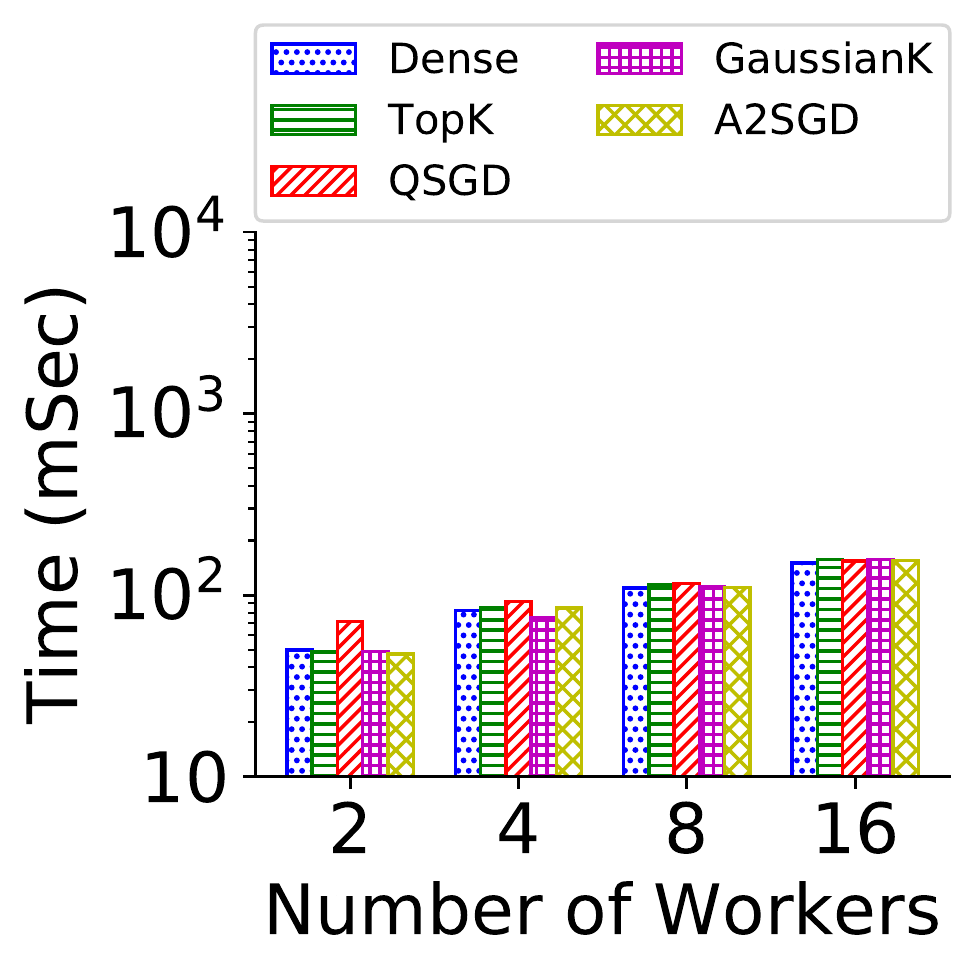}
	\caption{FNN3}
	\label{fig:fnn3_speedup}
\end{subfigure}%
~ 
\begin{subfigure}[b]{0.25\textwidth}
	\centering
	\includegraphics[width=\linewidth]{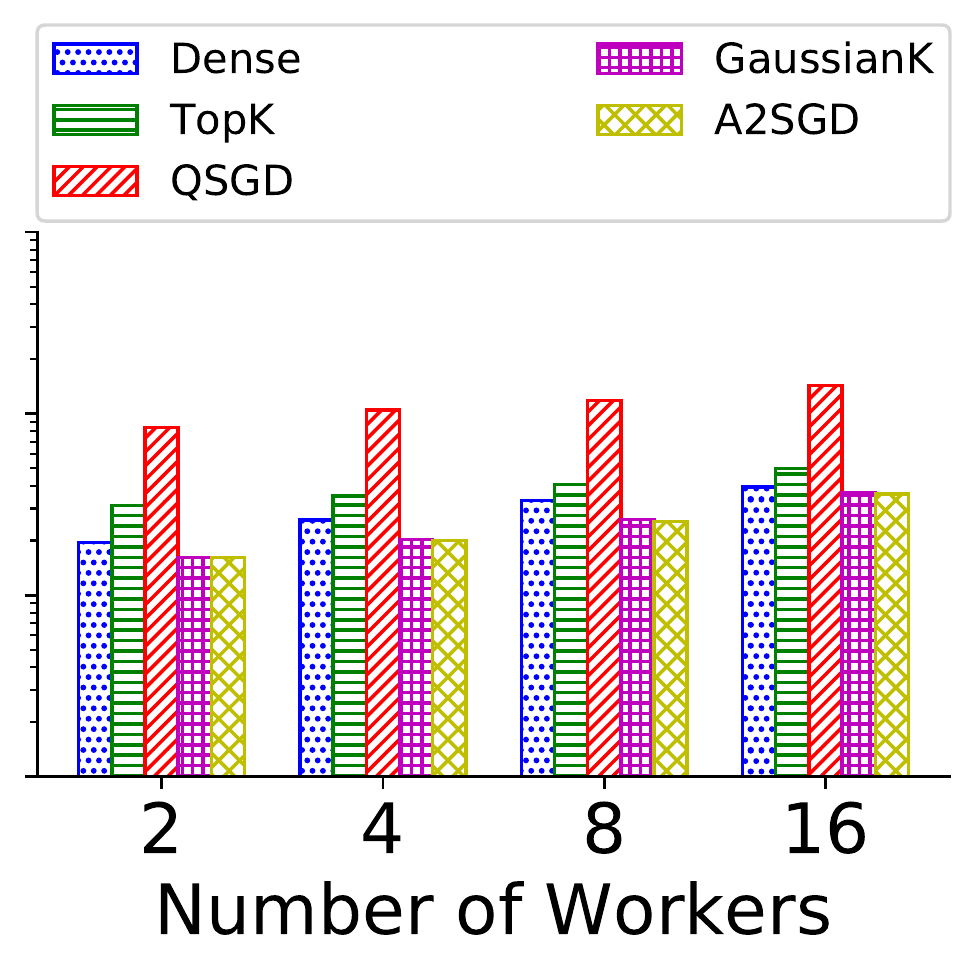}
	\caption{VGG16}
	\label{fig:vgg16_speedup}
\end{subfigure}%
~ 
\begin{subfigure}[b]{0.25\textwidth}
	\centering
	\includegraphics[width=\linewidth]{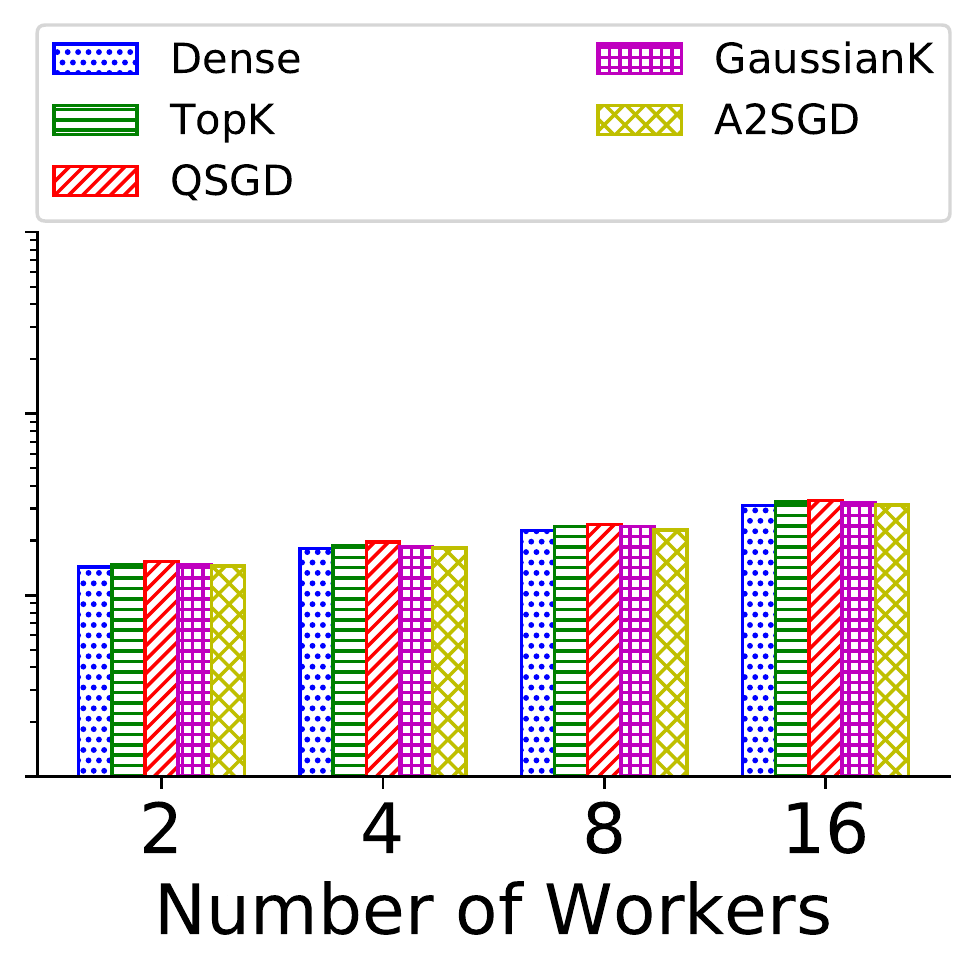}
	\caption{ResNet-20}
	\label{fig:resnet20_speedup}
\end{subfigure}%
~ 
\begin{subfigure}[b]{0.25\textwidth}
	\centering
	\includegraphics[width=\linewidth]{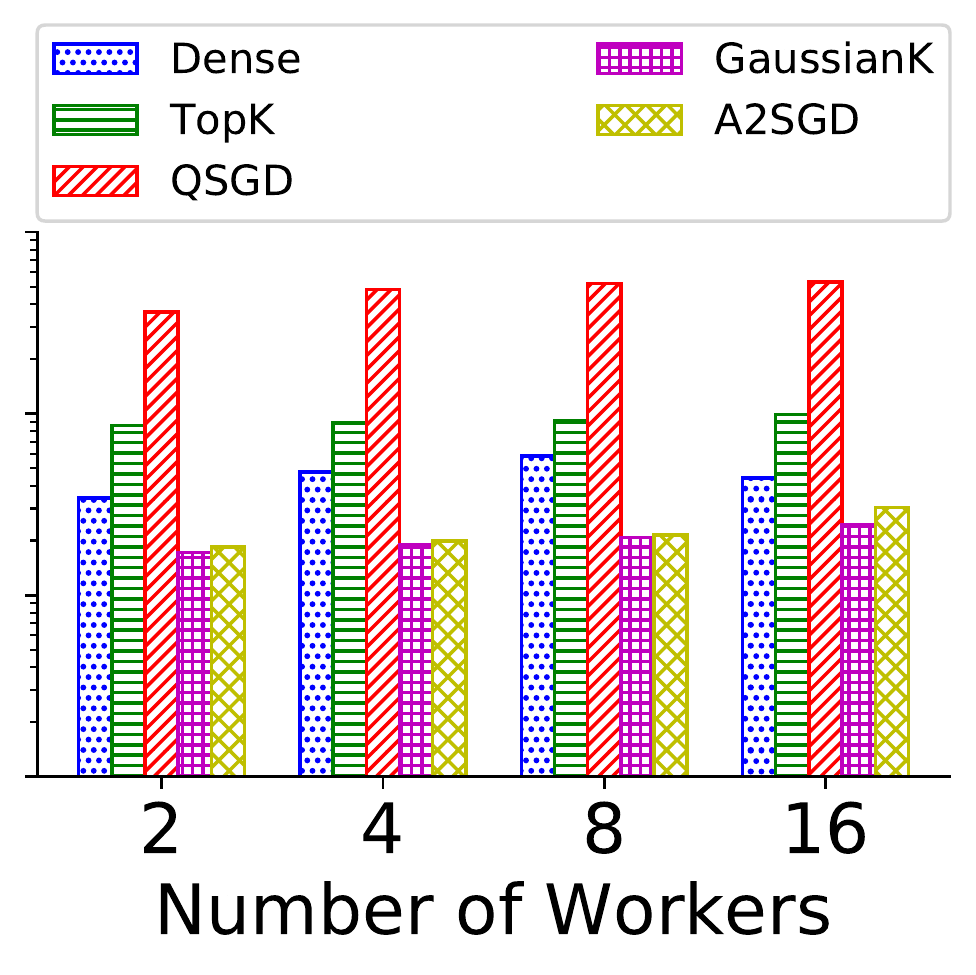}
	\caption{LSTM}
	\label{fig:lstm_time}
\end{subfigure}\\
	\caption{Comparison of Average Iteration Time}\label{fig:iteration}
\vspace{-1.0pc}
\end{figure}

Given our understanding on the complexity of A2SGD and its scaling efficiency,
we further evaluate the benefit of A2SGD to the execution time of DNN training
models. We first measure the average execution time per iteration for all algorithms.
Figure~\ref{fig:iteration} shows the comparison across four different models.
For the smaller models FNN-3 and ResNet-20, A2SGD and Gaussian-K perform comparably to dense SGD, 
and slightly better than QSGD and Top-K. These two
models have a smaller number of parameters, which 
lead to an immaterial difference 
on the 100-Gbps high-bandwidth network.
The longer execution time per iteration of QSGD and Top-K 
is due to their higher computation costs.

For the bigger models VGG-16 and LSTM-PTB, A2SGD and Gaussian-K deliver much
faster execution time per iteration, compared to dense SGD, Top-K and
QSGD. For the biggest model, Gaussian-K achieves better execution time per iteration than A2SGD. 
We could not attribute this difference to the comparisons listed in Table~\ref{tab:complexity}. 
By examining the implementation of Gaussian-K, 
we realize that this is because Gaussian-K uses Allgather for exchanging gradients, 
which is faster than Allreduce adopted by A2SGD on 100-Gbps high-bandwidth
networks~\cite{ben2018torsten,thakur2005optimization}.
Furthermore, the execution per iteration is always the longest for QSGD.
Compared to dense SGD, this is because the overhead from its high computation 
dominates over the benefit of communication reduction 
on the 100-Gbps high-bandwidth network. Moreover, all algorithms 
exhibit longer execution time per iteration with an increasing number of
workers. This is because of the collective nature of 
gradient synchronization, i.e., more communication time 
for synchronization across more workers.

\begin{figure}[H]
\vspace{-0.5pc}
	\centering
	\begin{subfigure}[b]{0.25\textwidth}
		\centering
		\includegraphics[width=\linewidth]{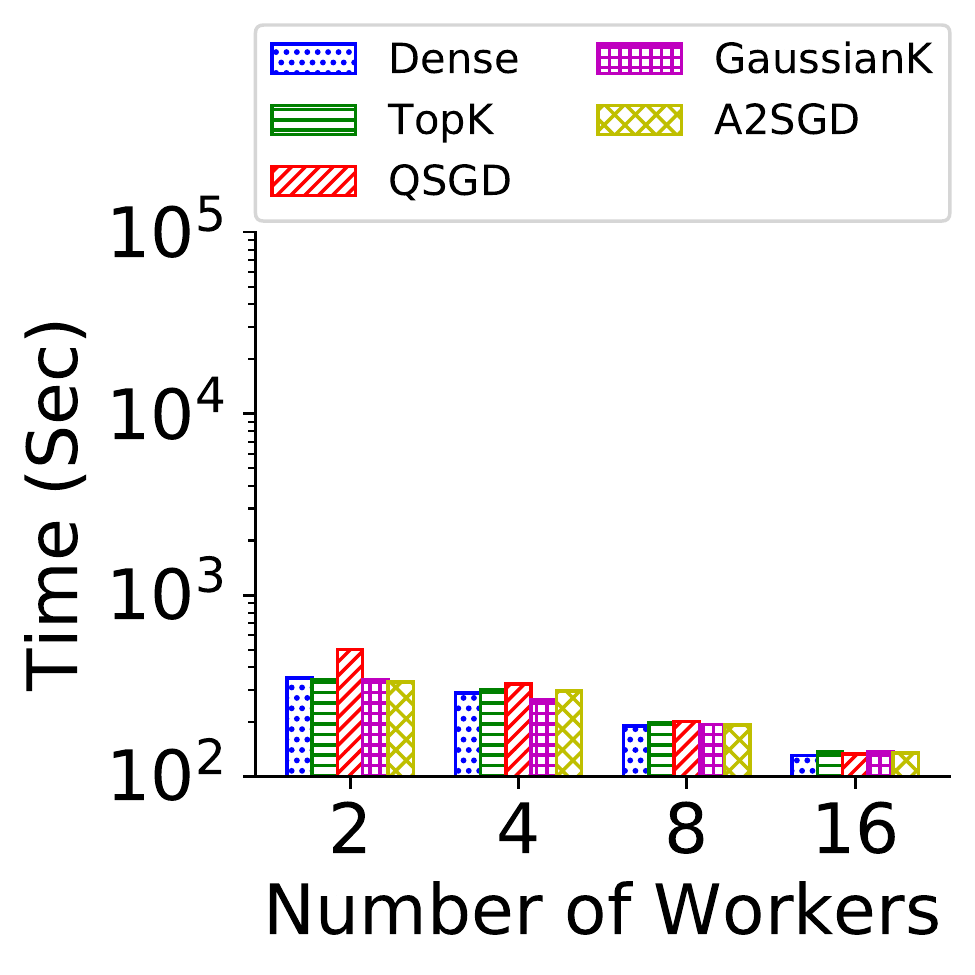}
		\caption{FNN3}
		\label{fig:fnn3_time}
	\end{subfigure}%
	~ 
	\begin{subfigure}[b]{0.25\textwidth}
		\centering
		\includegraphics[width=\linewidth]{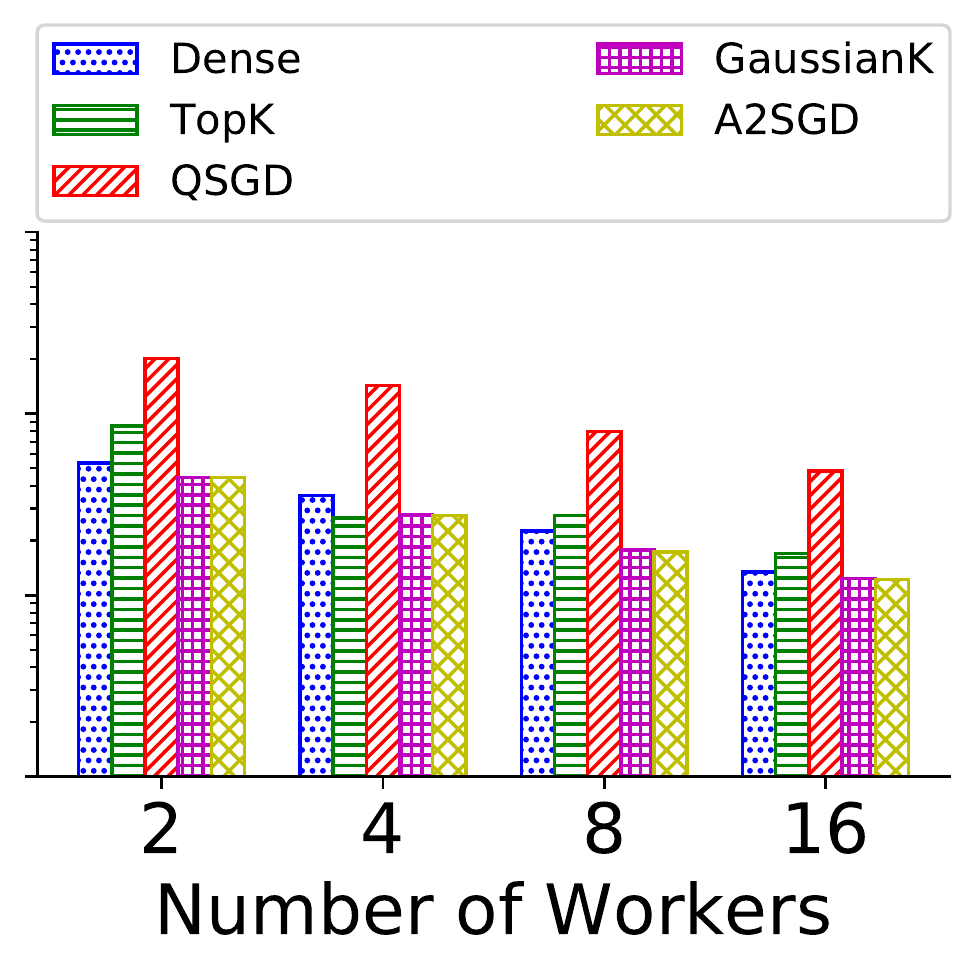}
		\caption{VGG16}
		\label{fig:vgg16_time}
	\end{subfigure}%
	~ 
	\begin{subfigure}[b]{0.25\textwidth}
		\centering
		\includegraphics[width=\linewidth]{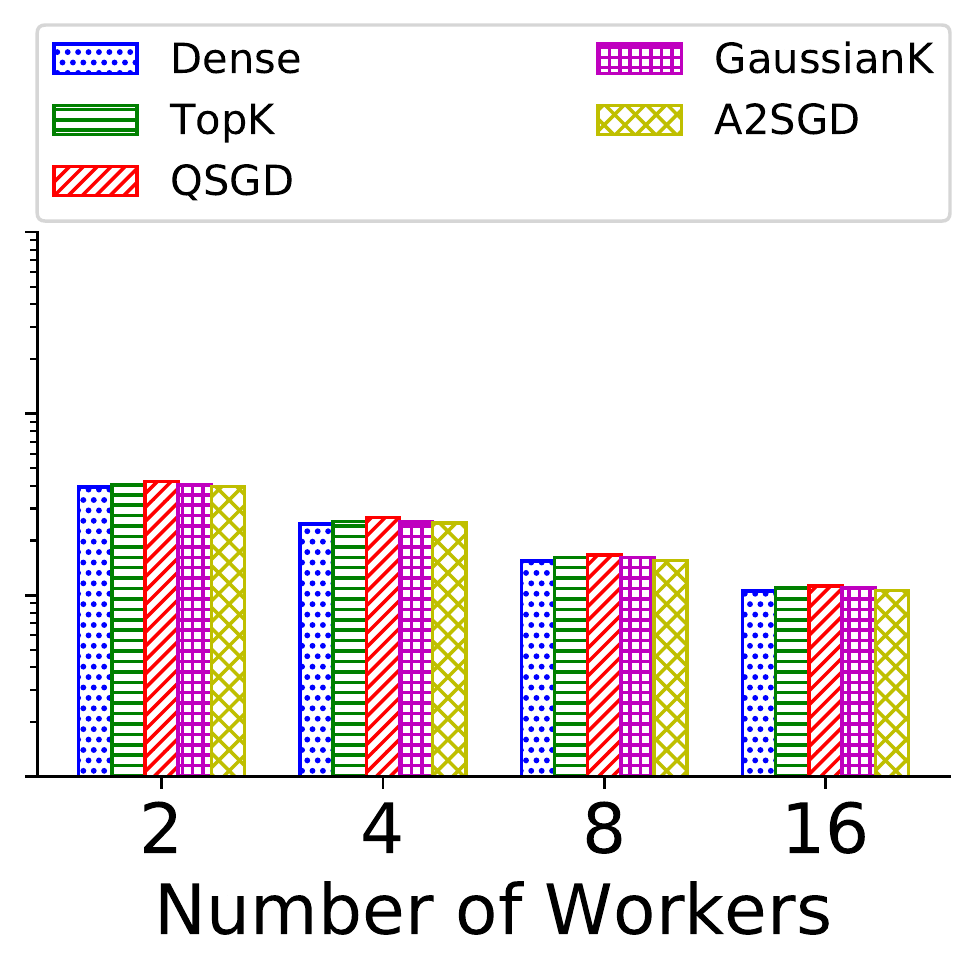}
		\caption{ResNet-20}
		\label{fig:resnet20_time}
	\end{subfigure}%
	~ 
	\begin{subfigure}[b]{0.25\textwidth}
		\centering
		\includegraphics[width=\linewidth]{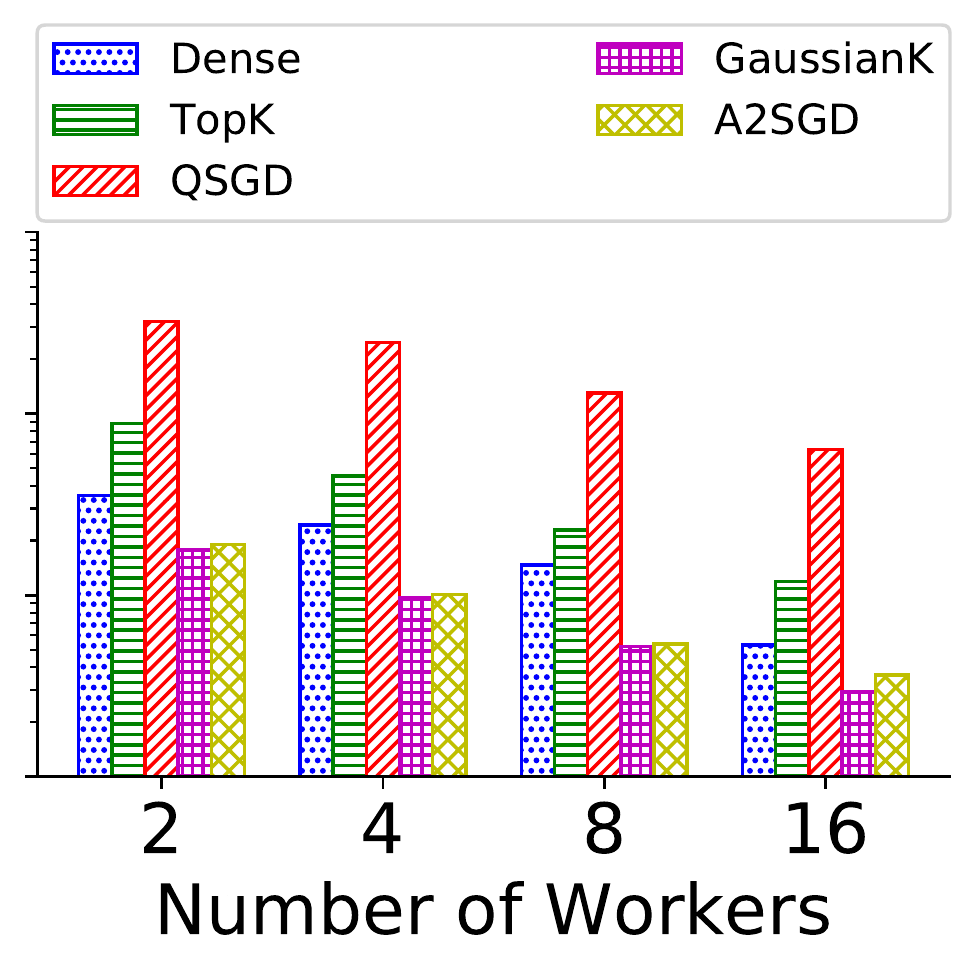}
		\caption{LSTM}
		\label{fig:lstm_time}
	\end{subfigure}\\
	\caption{Comparison of Total Execution Time}\label{fig:total}
\vspace{-1.0pc}
\end{figure}

We also evaluate the benefit of A2SGD to the total execution time with an increasing number of workers. 
Figure~\ref{fig:total} shows the comparison across all algorithms for four different models.
Despite the increasing execution time per iteration, all algorithms deliver
faster total execution time with more workers, a manifestation on the strength
of data parallel distributed SGD algorithms. 
Again, for FNN-3 and ResNet-20, A2SGD and Gaussian-K perform comparably to dense SGD, 
and slightly better than QSGD and Top-K, for the same reasons as previously stated.
For the bigger models VGG-16 and LSTM-PTB, A2SGD and Gaussian-K again 
achieve better performance than dense SGD, Top-K and QSGD. 
Gaussian-K is slightly faster than A2SGD for its Allgather implementation.
QSGD suffers from its high computation overhead compared to dense SGD, 
and its high communication costs compared to the other models.

\paragraph{Discussion.}
These evaluation results demonstrate that,
while achieving $\Or(1)$ communication complexity,
A2SGD delivers the best overall performance in terms of convergence accuracy,
scaling efficiency and execution time. Our comparative analysis
also identifies an opportunity for A2SGD.
While we have adopted the Allreduce implementation for gradient exchange in 
our initial implementation of A2SGD, it can be expanded with an 
Allgather-based alternative like Gaussian-K. 
We plan to pursue this optimization and update all our evaluation results. 
Also note that, compared to Gaussian-K, A2SGD does not have to go through 
an initial estimation of the threshold, and its computation cost is slightly lower.

\section{Conclusion}

In this paper, we have examined the scalability challenge of gradient
synchronization in distributed SGD and proposed a two-level gradient averaging
algorithm A2SGD for distributed workers to exchange only two averages 
and achieve $\Or(1)$ communication complexity per worker. 
We have theoretically analyzed the convergence of A2SGD. Our
experimental results have confirmed the convergence of A2SGD and demonstrated
that A2SGD achieves an overall improvement compared to other sparsification and
quantization algorithms~\cite{shi2019_GaussianTopK,stich2018sparsified,alistarh2017qsgd}. 
For all these algorithms, we also have systematically analyzed their 
computation and communication complexities during gradient synchronization
and pointed out that A2SGD outperforms the others asymptotically.


\begin{ack}
	We would like to thank Ms. Yue Zhu from the Computer Architecture and 
	SysTems Research Lab (CASTL) of Florida State University for her help on 
	the initial experimental setup and valuable suggestions related to this 
	work.
	This work used the Extreme Science and Engineering Discovery Environment 
	(XSEDE~\cite{towns2014xsede}), which is supported by National Science 
	Foundation grant number ACI-1548562.
	
	This work is also supported in part by the National Science Foundation 
	awards 1561041, 1564647, 1763547, and 1822737. Any opinions, findings, and 
	conclusions or recommendations expressed in this material are those of the 
	authors and do not necessarily reflect the views of the National Science 
	Foundation.
\end{ack}

%

\small{
\bibliographystyle{unsrtnat}
\bibliography{neurips_2020,mldl}

\begin{thebibliography}{51}
\providecommand{\natexlab}[1]{#1}
\providecommand{\url}[1]{\texttt{#1}}
\expandafter\ifx\csname urlstyle\endcsname\relax
  \providecommand{\doi}[1]{doi: #1}\else
  \providecommand{\doi}{doi: \begingroup \urlstyle{rm}\Url}\fi

\bibitem[Szegedy et~al.(2015)Szegedy, Liu, Jia, Sermanet, Reed, Anguelov,
  Erhan, Vanhoucke, Rabinovich, et~al.]{szegedy2015going}
Christian Szegedy, Wei Liu, Yangqing Jia, Pierre Sermanet, Scott Reed, Dragomir
  Anguelov, Dumitru Erhan, Vincent Vanhoucke, Andrew Rabinovich, et~al.
\newblock {Going Deeper with Convolutions}.
\newblock In \emph{2015 IEEE Conference on Computer Vision and Pattern
  Recognition (CVPR)}. Cvpr, 2015.

\bibitem[He et~al.(2016)He, Zhang, Ren, and Sun]{resnet1k}
Kaiming He, Xiangyu Zhang, Shaoqing Ren, and Jian Sun.
\newblock {Identity Mappings in Deep Residual Networks}.
\newblock \emph{CoRR}, absscaffe,/1603.05027, 2016.
\newblock URL \url{http://arxiv.org/abs/1603.05027}.

\bibitem[Radford et~al.(2019)Radford, Wu, Child, Luan, Amodei, and
  Sutskever]{GPT2}
Alec Radford, Jeffrey Wu, Rewon Child, David Luan, Dario Amodei, and Ilya
  Sutskever.
\newblock Language models are unsupervised multitask learners.
\newblock \emph{OpenAI Blog}, 1\penalty0 (8):\penalty0 9, 2019.

\bibitem[Vaswani et~al.(2017)Vaswani, Shazeer, Parmar, Uszkoreit, Jones, Gomez,
  Kaiser, and Polosukhin]{Transformer6B}
Ashish Vaswani, Noam Shazeer, Niki Parmar, Jakob Uszkoreit, Llion Jones,
  Aidan~N Gomez, \L~ukasz Kaiser, and Illia Polosukhin.
\newblock Attention is all you need.
\newblock In I.~Guyon, U.~V. Luxburg, S.~Bengio, H.~Wallach, R.~Fergus,
  S.~Vishwanathan, and R.~Garnett, editors, \emph{Advances in Neural
  Information Processing Systems 30}, pages 5998--6008. Curran Associates,
  Inc., 2017.
\newblock URL
  \url{http://papers.nips.cc/paper/7181-attention-is-all-you-need.pdf}.

\bibitem[Agarwal and Duchi(2011)]{agarwal2011distributed}
Alekh Agarwal and John~C. Duchi.
\newblock Distributed delayed stochastic optimization, 2011.

\bibitem[Strom(2015)]{strom2015scalable}
Nikko Strom.
\newblock Scalable distributed dnn training using commodity gpu cloud
  computing.
\newblock In \emph{Sixteenth Annual Conference of the International Speech
  Communication Association}, 2015.

\bibitem[Ho et~al.(2013)Ho, Cipar, Cui, Kim, Lee, Gibbons, Gibson, Ganger, and
  Xing]{NIPS13SML}
Qirong Ho, James Cipar, Henggang Cui, Jin~Kyu Kim, Seunghak Lee, Phillip~B.
  Gibbons, Garth~A. Gibson, Gregory~R. Ganger, and Eric~P. Xing.
\newblock More effective distributed ml via a stale synchronous parallel
  parameter server.
\newblock In \emph{Proceedings of the 26th International Conference on Neural
  Information Processing Systems - Volume 1}, NIPS’13, page 1223–1231, Red
  Hook, NY, USA, 2013. Curran Associates Inc.

\bibitem[Li et~al.(2014{\natexlab{a}})Li, Andersen, Park, Smola, Ahmed,
  Josifovski, Long, Shekita, and Su]{OSDI14SDML}
Mu~Li, David~G. Andersen, Jun~Woo Park, Alexander~J. Smola, Amr Ahmed, Vanja
  Josifovski, James Long, Eugene~J. Shekita, and Bor-Yiing Su.
\newblock Scaling distributed machine learning with the parameter server.
\newblock In \emph{Proceedings of the 11th USENIX Conference on Operating
  Systems Design and Implementation}, OSDI’14, page 583–598, USA,
  2014{\natexlab{a}}. USENIX Association.
\newblock ISBN 9781931971164.

\bibitem[Chilimbi et~al.(2014)Chilimbi, Suzue, Apacible, and
  Kalyanaraman]{OSDI14Adam}
Trishul Chilimbi, Yutaka Suzue, Johnson Apacible, and Karthik Kalyanaraman.
\newblock Project adam: Building an efficient and scalable deep learning
  training system.
\newblock In \emph{11th {USENIX} Symposium on Operating Systems Design and
  Implementation ({OSDI} 14)}, pages 571--582, Broomfield, CO, October 2014.
  {USENIX} Association.
\newblock ISBN 978-1-931971-16-4.
\newblock URL
  \url{https://www.usenix.org/conference/osdi14/technical-sessions/presentation/chilimbi}.

\bibitem[Goyal et~al.(2017)Goyal, Doll{\'a}r, Girshick, Noordhuis, Wesolowski,
  Kyrola, Tulloch, Jia, and He]{goyal2017accurate}
Priya Goyal, Piotr Doll{\'a}r, Ross Girshick, Pieter Noordhuis, Lukasz
  Wesolowski, Aapo Kyrola, Andrew Tulloch, Yangqing Jia, and Kaiming He.
\newblock {Accurate, Large Minibatch SGD: Training Imagenet in 1 Hour}.
\newblock \emph{arXiv preprint arXiv:1706.02677}, 2017.

\bibitem[You et~al.(2017)You, Gitman, and Ginsburg]{you2017scaling}
Yang You, Igor Gitman, and Boris Ginsburg.
\newblock {Scaling SGD batch size to 32k for Imagenet Training}.
\newblock \emph{arXiv preprint arXiv:1708.03888}, 2017.

\bibitem[Li et~al.(2014{\natexlab{b}})Li, Zhang, Chen, and
  Smola]{li2014efficient}
Mu~Li, Tong Zhang, Yuqiang Chen, and Alexander~J Smola.
\newblock {Efficient Mini-batch Training for Stochastic Optimization}.
\newblock In \emph{Proceedings of the 20th ACM SIGKDD international conference
  on Knowledge discovery and data mining}, pages 661--670. ACM,
  2014{\natexlab{b}}.

\bibitem[Gibiansky and Hestness(2017)]{baiduallreduce}
Andrew Gibiansky and Joel Hestness.
\newblock {Baidu Research, TensorFlow-Allreduce}.
\newblock {https://github.com/baidu-research/tensorflow-allreduce}, 2017.

\bibitem[{Awan} et~al.(2019){Awan}, {Bédorf}, {Chu}, {Subramoni}, and
  {Panda}]{awan:ccgrid19}
A.~A. {Awan}, J.~{Bédorf}, C.~{Chu}, H.~{Subramoni}, and D.~K. {Panda}.
\newblock Scalable distributed dnn training using tensorflow and cuda-aware
  mpi: Characterization, designs, and performance evaluation.
\newblock In \emph{2019 19th IEEE/ACM International Symposium on Cluster, Cloud
  and Grid Computing (CCGRID)}, pages 498--507, May 2019.
\newblock \doi{10.1109/CCGRID.2019.00064}.

\bibitem[Jiang and Agrawal(2018)]{jiang2018linear}
Peng Jiang and Gagan Agrawal.
\newblock A linear speedup analysis of distributed deep learning with sparse
  and quantized communication.
\newblock In \emph{Advances in Neural Information Processing Systems}, pages
  2525--2536, 2018.

\bibitem[LeCun et~al.(1990)LeCun, Denker, and Solla]{NIPS1989_OBD}
Yann LeCun, John~S. Denker, and Sara~A. Solla.
\newblock Optimal brain damage.
\newblock In D.~S. Touretzky, editor, \emph{Advances in Neural Information
  Processing Systems 2}, pages 598--605. Morgan-Kaufmann, 1990.
\newblock URL \url{http://papers.nips.cc/paper/250-optimal-brain-damage.pdf}.

\bibitem[Hassibi and Stork(1993)]{NIPS1992_OBS}
Babak Hassibi and David~G. Stork.
\newblock Second order derivatives for network pruning: Optimal brain surgeon.
\newblock In S.~J. Hanson, J.~D. Cowan, and C.~L. Giles, editors,
  \emph{Advances in Neural Information Processing Systems 5}, pages 164--171.
  Morgan-Kaufmann, 1993.
\newblock URL
  \url{http://papers.nips.cc/paper/647-second-order-derivatives-for-network-pruning-optimal-brain-surgeon.pdf}.

\bibitem[Guo et~al.(2016)Guo, Yao, and Chen]{NIPS2016_DNS}
Yiwen Guo, Anbang Yao, and Yurong Chen.
\newblock Dynamic network surgery for efficient dnns.
\newblock In D.~D. Lee, M.~Sugiyama, U.~V. Luxburg, I.~Guyon, and R.~Garnett,
  editors, \emph{Advances in Neural Information Processing Systems 29}, pages
  1379--1387. Curran Associates, Inc., 2016.
\newblock URL
  \url{http://papers.nips.cc/paper/6165-dynamic-network-surgery-for-efficient-dnns.pdf}.

\bibitem[Han et~al.(2016)Han, Mao, and Dally]{ICLR2016DC}
Song Han, Huizi Mao, and William~J. Dally.
\newblock Deep compression: Compressing deep neural network with pruning,
  trained quantization and huffman coding.
\newblock In Yoshua Bengio and Yann LeCun, editors, \emph{4th International
  Conference on Learning Representations, {ICLR} 2016, San Juan, Puerto Rico,
  May 2-4, 2016, Conference Track Proceedings}, 2016.
\newblock URL \url{http://arxiv.org/abs/1510.00149}.

\bibitem[Wen et~al.(2017)Wen, Xu, Yan, Wu, Wang, Chen, and Li]{wen2017terngrad}
Wei Wen, Cong Xu, Feng Yan, Chunpeng Wu, Yandan Wang, Yiran Chen, and Hai Li.
\newblock Terngrad: Ternary gradients to reduce communication in distributed
  deep learning.
\newblock In \emph{Advances in neural information processing systems}, pages
  1509--1519, 2017.

\bibitem[Alistarh et~al.(2017)Alistarh, Grubic, Li, Tomioka, and
  Vojnovic]{alistarh2017qsgd}
Dan Alistarh, Demjan Grubic, Jerry Li, Ryota Tomioka, and Milan Vojnovic.
\newblock Qsgd: Communication-efficient sgd via gradient quantization and
  encoding.
\newblock In \emph{Advances in Neural Information Processing Systems}, pages
  1709--1720, 2017.

\bibitem[Karimireddy et~al.(2019)Karimireddy, Rebjock, Stich, and
  Jaggi]{karimireddy2019error}
Sai~Praneeth Karimireddy, Quentin Rebjock, Sebastian~U Stich, and Martin Jaggi.
\newblock Error feedback fixes signsgd and other gradient compression schemes.
\newblock \emph{arXiv preprint arXiv:1901.09847}, 2019.

\bibitem[Bernstein et~al.(2018)Bernstein, Zhao, Azizzadenesheli, and
  Anandkumar]{bernstein2018signsgd}
Jeremy Bernstein, Jiawei Zhao, Kamyar Azizzadenesheli, and Anima Anandkumar.
\newblock signsgd with majority vote is communication efficient and fault
  tolerant, 2018.

\bibitem[Alistarh et~al.(2018)Alistarh, Hoefler, Johansson, Konstantinov,
  Khirirat, and Renggli]{alistarh2018convergence}
Dan Alistarh, Torsten Hoefler, Mikael Johansson, Nikola Konstantinov, Sarit
  Khirirat, and C{\'e}dric Renggli.
\newblock The convergence of sparsified gradient methods.
\newblock In \emph{Advances in Neural Information Processing Systems}, pages
  5973--5983, 2018.

\bibitem[Shi et~al.(2019{\natexlab{a}})Shi, Chu, Cheung, and
  See]{shi2019_GaussianTopK}
Shaohuai Shi, Xiaowen Chu, Ka~Chun Cheung, and Simon See.
\newblock Understanding top-k sparsification in distributed deep learning.
\newblock \emph{arXiv preprint arXiv:1911.08772}, 2019{\natexlab{a}}.

\bibitem[Aji and Heafield(2017)]{aji-heafield-2017-sparse}
Alham~Fikri Aji and Kenneth Heafield.
\newblock Sparse communication for distributed gradient descent.
\newblock In \emph{Proceedings of the 2017 Conference on Empirical Methods in
  Natural Language Processing}, pages 440--445, Copenhagen, Denmark, September
  2017. Association for Computational Linguistics.
\newblock \doi{10.18653/v1/D17-1045}.
\newblock URL \url{https://www.aclweb.org/anthology/D17-1045}.

\bibitem[Stich et~al.(2018)Stich, Cordonnier, and Jaggi]{stich2018sparsified}
Sebastian~U Stich, Jean-Baptiste Cordonnier, and Martin Jaggi.
\newblock Sparsified sgd with memory.
\newblock In \emph{Advances in Neural Information Processing Systems}, pages
  4447--4458, 2018.

\bibitem[Seide et~al.(2014)Seide, Fu, Droppo, Li, and Yu]{seide20141}
Frank Seide, Hao Fu, Jasha Droppo, Gang Li, and Dong Yu.
\newblock 1-bit stochastic gradient descent and its application to
  data-parallel distributed training of speech dnns.
\newblock In \emph{Fifteenth Annual Conference of the International Speech
  Communication Association}, 2014.

\bibitem[Wu et~al.(2018)Wu, Huang, Huang, and Zhang]{wu2018error}
Jiaxiang Wu, Weidong Huang, Junzhou Huang, and Tong Zhang.
\newblock Error compensated quantized sgd and its applications to large-scale
  distributed optimization.
\newblock \emph{arXiv preprint arXiv:1806.08054}, 2018.

\bibitem[Tang et~al.(2019)Tang, Lian, Zhang, and Liu]{tang2019doublesqueeze}
Hanlin Tang, Xiangru Lian, Tong Zhang, and Ji~Liu.
\newblock Doublesqueeze: Parallel stochastic gradient descent with double-pass
  error-compensated compression.
\newblock \emph{arXiv preprint arXiv:1905.05957}, 2019.

\bibitem[Haddadpour et~al.(2019)Haddadpour, Kamani, Mahdavi, and
  Cadambe]{haddadpour2019trading}
Farzin Haddadpour, Mohammad~Mahdi Kamani, Mehrdad Mahdavi, and Viveck Cadambe.
\newblock Trading redundancy for communication: Speeding up distributed sgd for
  non-convex optimization.
\newblock In \emph{International Conference on Machine Learning}, pages
  2545--2554, 2019.

\bibitem[K\"{o}ster et~al.(2017)K\"{o}ster, Webb, Wang, Nassar, Bansal,
  Constable, Elibol, Gray, Hall, Hornof, Khosrowshahi, Kloss, Pai, and
  Rao]{NIPS2017_flexpoint}
Urs K\"{o}ster, Tristan Webb, Xin Wang, Marcel Nassar, Arjun~K Bansal, William
  Constable, Oguz Elibol, Scott Gray, Stewart Hall, Luke Hornof, Amir
  Khosrowshahi, Carey Kloss, Ruby~J Pai, and Naveen Rao.
\newblock Flexpoint: An adaptive numerical format for efficient training of
  deep neural networks.
\newblock In I.~Guyon, U.~V. Luxburg, S.~Bengio, H.~Wallach, R.~Fergus,
  S.~Vishwanathan, and R.~Garnett, editors, \emph{Advances in Neural
  Information Processing Systems 30}, pages 1742--1752. Curran Associates,
  Inc., 2017.

\bibitem[Narang et~al.(2018)Narang, Diamos, Elsen, Micikevicius, Alben, Garcia,
  Ginsburg, Houston, Kuchaiev, Venkatesh, et~al.]{narang2018mixed}
Sharan Narang, Gregory Diamos, Erich Elsen, Paulius Micikevicius, Jonah Alben,
  David Garcia, Boris Ginsburg, Michael Houston, Oleksii Kuchaiev, Ganesh
  Venkatesh, et~al.
\newblock Mixed precision training.
\newblock In \emph{Proc. 6th Int. Conf. on Learning Representations (ICLR)},
  2018.

\bibitem[Jia et~al.(2018)Jia, Song, He, Wang, Rong, Zhou, Xie, Guo, Yang, Yu,
  et~al.]{jia2018highly}
Xianyan Jia, Shutao Song, Wei He, Yangzihao Wang, Haidong Rong, Feihu Zhou,
  Liqiang Xie, Zhenyu Guo, Yuanzhou Yang, Liwei Yu, et~al.
\newblock Highly scalable deep learning training system with mixed-precision:
  Training imagenet in four minutes.
\newblock \emph{arXiv preprint arXiv:1807.11205}, 2018.

\bibitem[Wang et~al.(2018)Wang, Choi, Brand, Chen, and
  Gopalakrishnan]{NIPS2018_wang8bit}
Naigang Wang, Jungwook Choi, Daniel Brand, Chia-Yu Chen, and Kailash
  Gopalakrishnan.
\newblock Training deep neural networks with 8-bit floating point numbers.
\newblock In S.~Bengio, H.~Wallach, H.~Larochelle, K.~Grauman, N.~Cesa-Bianchi,
  and R.~Garnett, editors, \emph{Advances in Neural Information Processing
  Systems 31}, pages 7675--7684. Curran Associates, Inc., 2018.
\newblock URL
  \url{http://papers.nips.cc/paper/7994-training-deep-neural-networks-with-8-bit-floating-point-numbers.pdf}.

\bibitem[Choi et~al.(2019)Choi, Venkataramani, Srinivasan, Gopalakrishnan,
  Wang, and Chuang]{Choi2019ACCURATEAE}
Jungwook Choi, Swagath Venkataramani, Vijayalakshmi Srinivasan, Kailash
  Gopalakrishnan, Zhuo Wang, and Pierce Chuang.
\newblock Accurate and efficient 2-bit quantized neural networks.
\newblock In \emph{The 2nd SysML Conference}, Palo Alto, CA, USA, 2019.

\bibitem[Lin et~al.(2017)Lin, Han, Mao, Wang, and Dally]{lin2017deep}
Yujun Lin, Song Han, Huizi Mao, Yu~Wang, and William~J Dally.
\newblock Deep gradient compression: Reducing the communication bandwidth for
  distributed training.
\newblock \emph{arXiv preprint arXiv:1712.01887}, 2017.

\bibitem[Wangni et~al.(2018)Wangni, Wang, Liu, and Zhang]{wangni2018gradient}
Jianqiao Wangni, Jialei Wang, Ji~Liu, and Tong Zhang.
\newblock Gradient sparsification for communication-efficient distributed
  optimization.
\newblock In \emph{Advances in Neural Information Processing Systems}, pages
  1299--1309, 2018.

\bibitem[Shi et~al.(2019{\natexlab{b}})Shi, Zhao, Wang, Tang, and
  Chu]{shi2019convergence}
Shaohuai Shi, Kaiyong Zhao, Qiang Wang, Zhenheng Tang, and Xiaowen Chu.
\newblock A convergence analysis of distributed sgd with
  communication-efficient gradient sparsification.
\newblock In \emph{Proceedings of the Twenty-Eighth International Joint
  Conference on Artificial Intelligence, IJCAI-19}, pages 3411--3417,
  2019{\natexlab{b}}.

\bibitem[Renggli et~al.(2019)Renggli, Ashkboos, Aghagolzadeh, Alistarh, and
  Hoefler]{SC19_SparCML}
Cedric Renggli, Saleh Ashkboos, Mehdi Aghagolzadeh, Dan Alistarh, and Torsten
  Hoefler.
\newblock Sparcml: High-performance sparse communication for machine learning.
\newblock In \emph{Proceedings of the International Conference for High
  Performance Computing, Networking, Storage and Analysis}, SC ’19, New York,
  NY, USA, 2019. Association for Computing Machinery.
\newblock ISBN 9781450362290.
\newblock \doi{10.1145/3295500.3356222}.
\newblock URL \url{https://doi.org/10.1145/3295500.3356222}.

\bibitem[{Adam Paszke and Sam Gross and Soumith Chintala and Gregory
  Chanan}({})]{PyTorch}
{Adam Paszke and Sam Gross and Soumith Chintala and Gregory Chanan}.
\newblock {Tensors and Dynamic neural networks in Python with strong GPU
  acceleration}.
\newblock {https://pytorch.org/}, {}.

\bibitem[git(2020{\natexlab{a}})]{git_qsgd}
{Sparsified SGD with Memory}.
\newblock https://github.com/epfml/sparsifiedSGD/blob/master/qsgd.py,
  2020{\natexlab{a}}.

\bibitem[Bottou(1998)]{bottou1998online}
L{\'e}on Bottou.
\newblock Online learning and stochastic approximations.
\newblock \emph{On-line learning in neural networks}, 17\penalty0 (9):\penalty0
  142, 1998.

\bibitem[cud({})]{cuda}
{CUDA}.
\newblock \url{https://developer.nvidia.com/cuda-10.1-download-archive-base},
  {}.

\bibitem[Sergeev and Del~Balso(2018)]{Horovod}
Alexander Sergeev and Mike Del~Balso.
\newblock Horovod: fast and easy distributed deep learning in tensorflow.
\newblock \emph{arXiv preprint arXiv:1802.05799}, 2018.

\bibitem[Thakur et~al.(2005)Thakur, Rabenseifner, and
  Gropp]{thakur2005optimization}
Rajeev Thakur, Rolf Rabenseifner, and William Gropp.
\newblock Optimization of collective communication operations in mpich.
\newblock \emph{The International Journal of High Performance Computing
  Applications}, 19\penalty0 (1):\penalty0 49--66, 2005.

\bibitem[git(2020{\natexlab{b}})]{git_gaussianK}
{Understanding Top-k Sparsification in Distributed Deep Learning}.
\newblock https://github.com/hclhkbu/GaussianK-SGD, 2020{\natexlab{b}}.

\bibitem[git(2016)]{git_topk}
{K-Selection Implementation of Torch TopK}.
\newblock https://github.com/torch/torch7/pull/496, 2016.

\bibitem[Shanbhag et~al.(2018)Shanbhag, Pirk, and Madden]{topk_query}
Anil Shanbhag, Holger Pirk, and Samuel Madden.
\newblock Efficient top-k query processing on massively parallel hardware.
\newblock In \emph{Proceedings of the 2018 International Conference on
  Management of Data}, SIGMOD ’18, page 1557–1570, New York, NY, USA, 2018.
  Association for Computing Machinery.
\newblock ISBN 9781450347037.
\newblock \doi{10.1145/3183713.3183735}.
\newblock URL \url{https://doi.org/10.1145/3183713.3183735}.

\bibitem[Ben-Nun(2018)]{ben2018torsten}
Tal Ben-Nun.
\newblock Torsten hoe er. 2018. demystifying parallel and distributed deep
  learning: An in-depth concurrency analysis.
\newblock \emph{arXiv preprint arXiv:1802.09941}, 2018.

\bibitem[Towns et~al.(2014)Towns, Cockerill, Dahan, Foster, Gaither, Grimshaw,
  Hazlewood, Lathrop, Lifka, Peterson, et~al.]{towns2014xsede}
John Towns, Timothy Cockerill, Maytal Dahan, Ian Foster, Kelly Gaither, Andrew
  Grimshaw, Victor Hazlewood, Scott Lathrop, Dave Lifka, Gregory~D Peterson,
  et~al.
\newblock Xsede: accelerating scientific discovery.
\newblock \emph{Computing in science \& engineering}, 16\penalty0 (5):\penalty0
  62--74, 2014.

\end{thebibliography}
}

\clearpage
\appendix

\roman{section}
\setcounter{section}{0}
\section{Appendix}

\subsection{Proof of Theorem 1 on the Convergence of A2SGD}
\label{subsec:proof}

\paragraph{Theorem 1.} When the learning system is updated as follows:
\begin{equation}
w_{t+1} = w_t - {\eta_t}(g_t + \nabla \mu_{t}), \label{eq:13}
\end{equation}
then, it \textbf{converges almost surely} toward minimum $w^*$, $i.e.\; 
\Prob(\lim\limits_{t\to+\infty}w_t = w^*) = 1$.

\paragraph{\textit\em\underline{Proof:}}

\begin{equation}
h_{t+1} - h_{t} = -2\eta_t(w-w^*)^T(g_t + \nabla \mu_{t}) 
+ \eta_t^2(g_t + \nabla \mu_{t}) \label{eq:14b}
\end{equation}

\textit{Taking conditional expectation of previous equation:}
\begin{equation}
\Expect{h_{t+1} - h_{t} | \mathcal{D}_t} = -2\eta_t(w_t - w^*)\Expect{g_t + \nabla \mu_{t} | \mathcal{D}_t} 
+ \eta_t^2\Expect{g_t + \nabla \mu_{t} | \mathcal{D}_t}
\label{eq:15}
\end{equation}
\textit{This expression can be further simplified using the following 
condition: $\nabla_wC(w_t) = (g_t + \nabla \mu_{t})$}
\begin{equation}
\Expect{h_{t+1} - h_{t} | \mathcal{D}_t} = -2\eta_t(w_t - 
w^*)\nabla_wC(w_t) 
+ \eta_t^2\Expect{\parallel g_t + \nabla \mu_{t} \parallel^2 | \mathcal{D}_t} \label{eq:16}
\end{equation}

\begin{equation}
\implies\Expect{h_{t+1} - h_{t} | \mathcal{D}_t} + 2\eta_t(w_t - 
w^*)\nabla_wC(w_t) =\eta_t^2\Expect{\parallel g_t + \nabla \mu_{t} \parallel^2 | \mathcal{D}_t} \label{eq:17b}
\end{equation}

\textit{From Assumption 3., we can further have:}
\begin{equation}
\Expect{h_{t+1} - h_{t} | \mathcal{D}_t} +2\eta_t(w_t - 
w^*)\nabla_wC(w_t)
\le \Alpha\eta_t^2 + \Beta\eta_t^2\parallel w-w^* \parallel^2
= \Alpha\eta_t^2 
+\Beta\eta_t^2h_t \label{eq:18}
\end{equation}
\begin{equation}
\Expect{h_{t+1} - (1+\eta_t^2\Beta)h_{t} | \mathcal{D}_t} \le -2\eta_t(w_t-w^*)^T\nabla_wC(w_t)+\eta_t^2\Alpha \label{eq:19}
\end{equation}
\textit{Eq.~\ref{eq:19} satisfies the condition of Lemma 1, which proves Theorem~\ref{eq:13}}.

\subsection{Convergence Accuracy with 2, 4 and 16 Workers}
\label{subsec:appConv}
\textbf{Note: \textit{All QSGD 
	results are evaluated using quantization level 4. Threshold for TopK and 
	GaussianK is 0.001d in all the experiments.}}

\begin{figure}[H]
	\centering
	\begin{subfigure}[b]{0.235\textwidth}
		\centering
		\includegraphics[width=\textwidth]{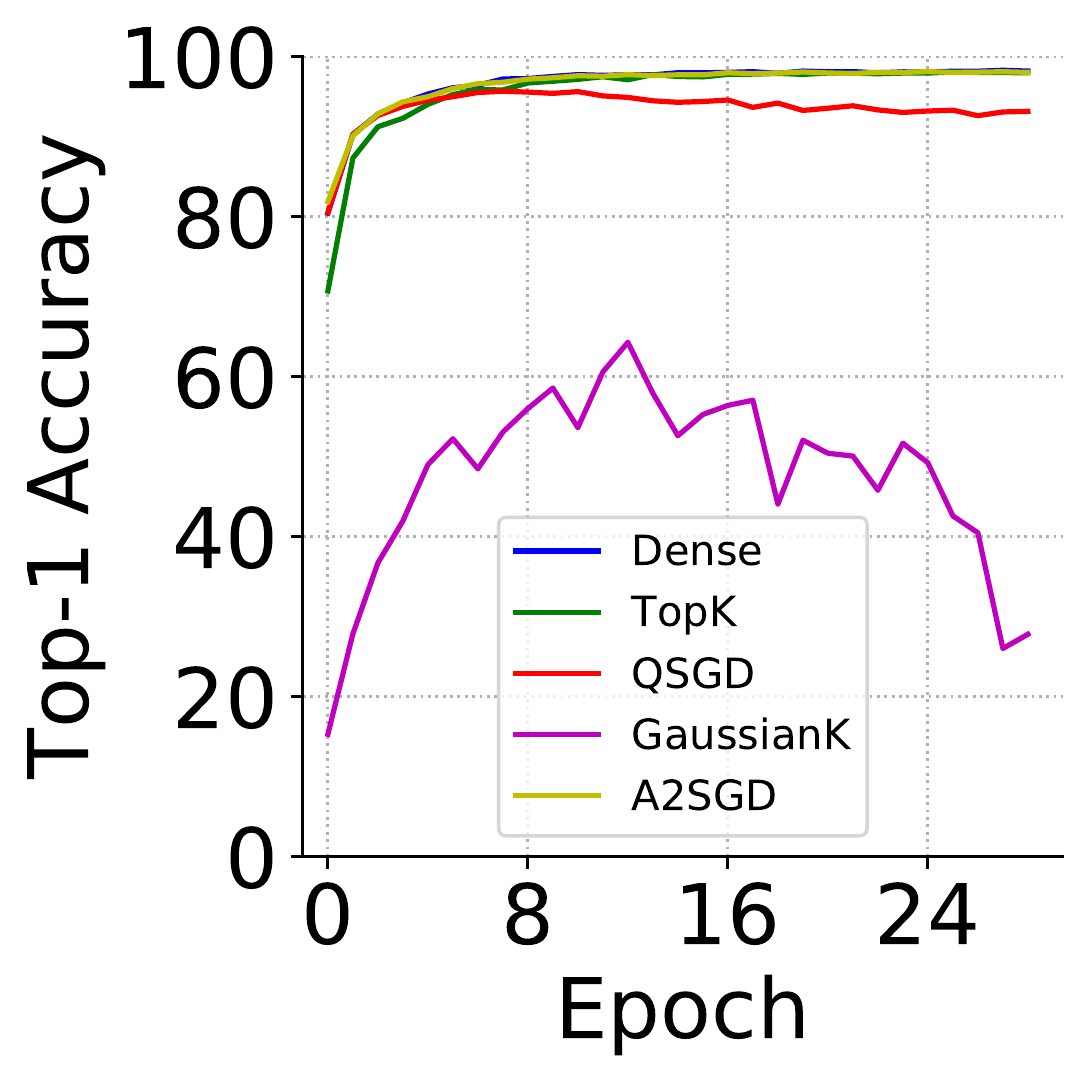}
		\caption{FNN3}
		\label{fig:FNN_acc}
	\end{subfigure}%
	~
	\begin{subfigure}[b]{0.215\textwidth}
		\centering
		\includegraphics[width=\textwidth]{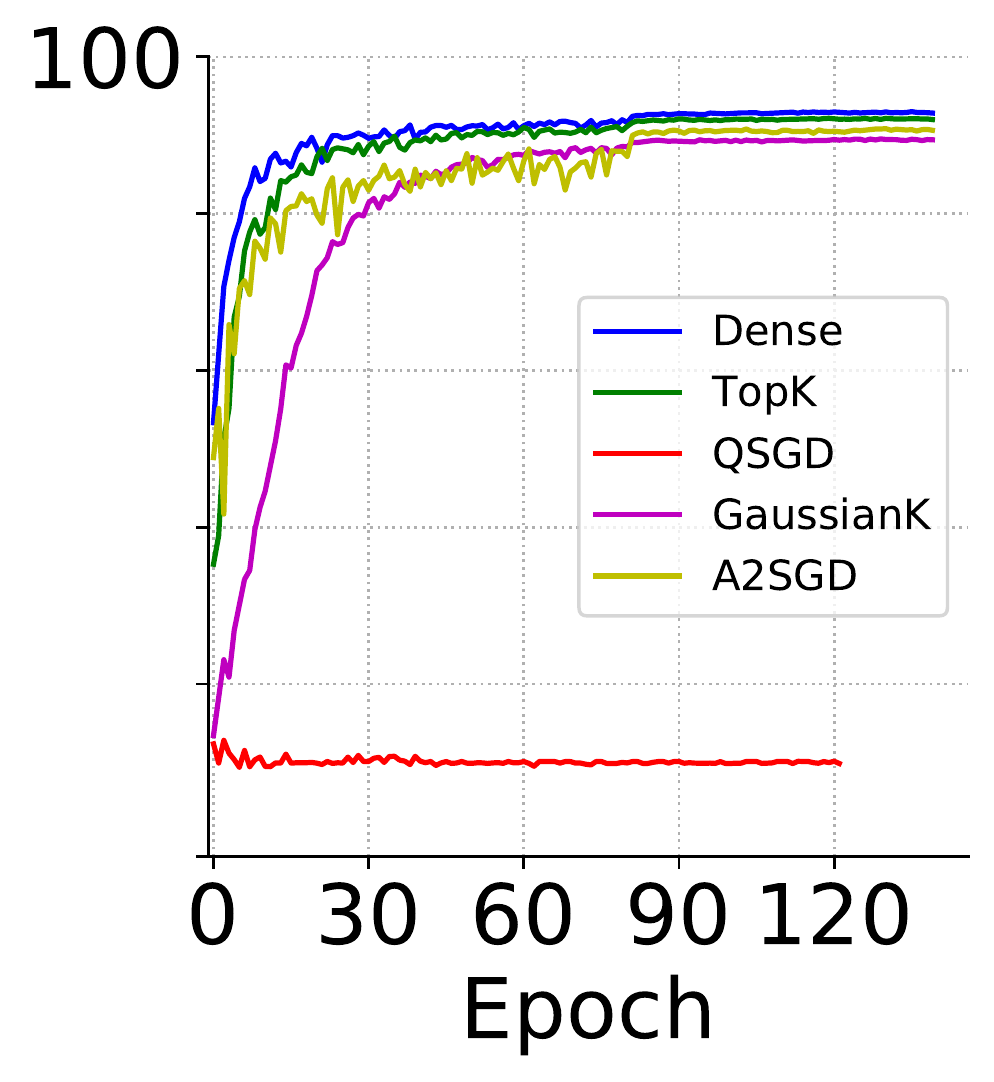}
		\caption{VGG16}
		\label{fig:VGG16_acc}
	\end{subfigure}%
	~
	\begin{subfigure}[b]{0.215\textwidth}
		\centering
		\includegraphics[width=\textwidth]{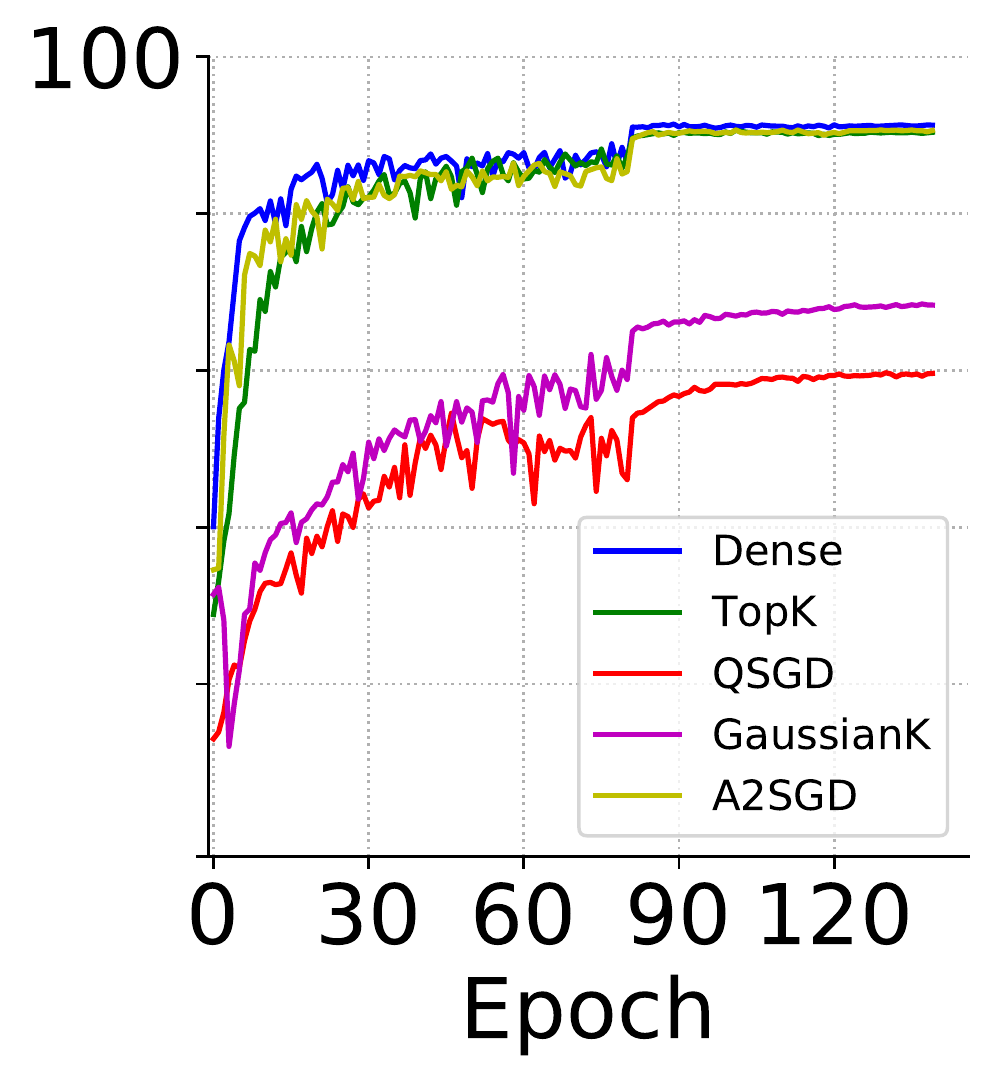}
		\caption{ResNet20}
		\label{fig:ResNet20_acc}
	\end{subfigure}%
	~
	\begin{subfigure}[b]{0.245\textwidth}
		\centering
		\includegraphics[width=\textwidth]{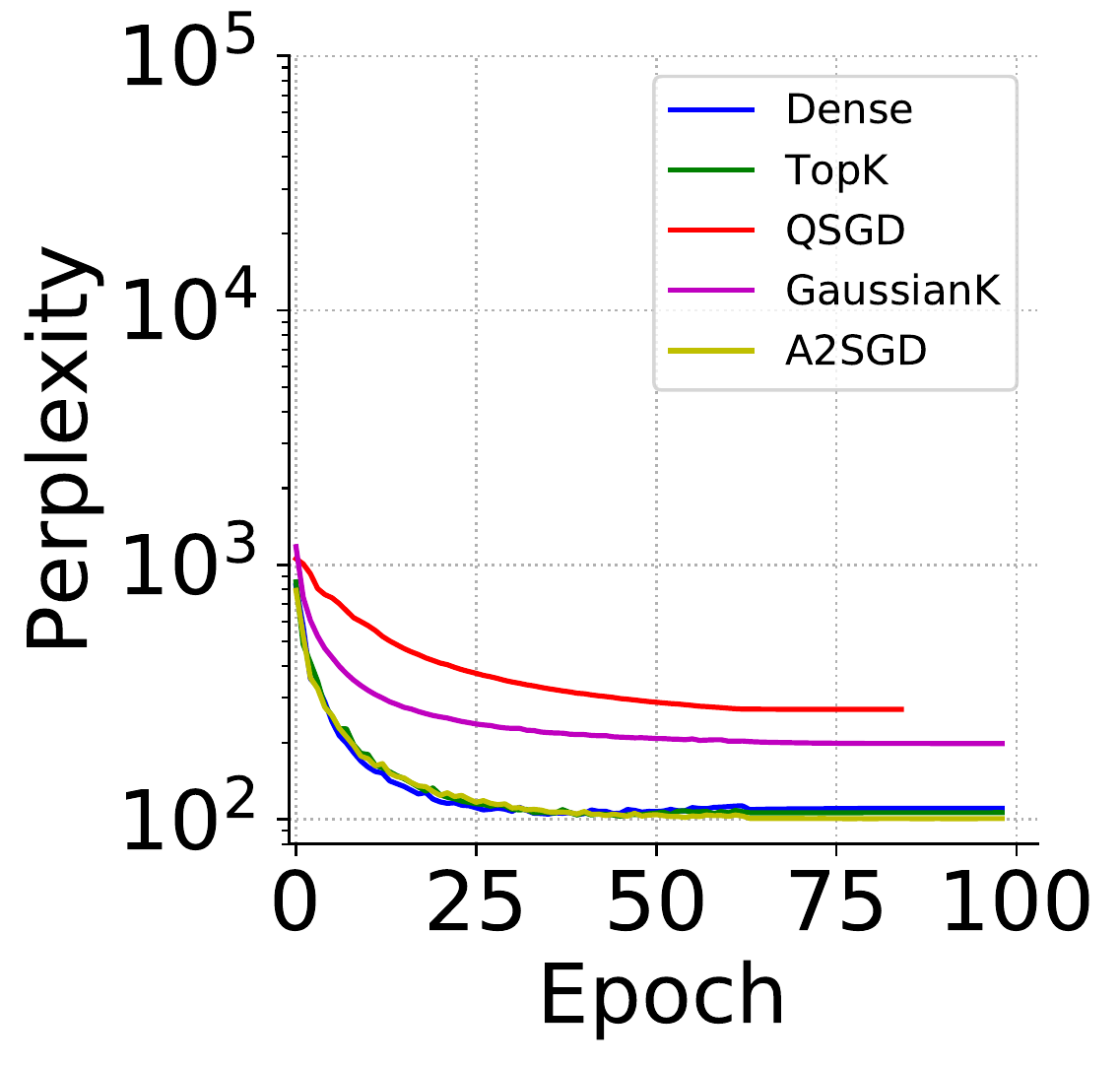}
		\caption{LSTM-PTB}
		\label{fig:LSTM_acc}
	\end{subfigure}
	\caption{Comparison of Convergence Accuracy with 2 Workers}
	\label{fig:conv2w}
\end{figure}

Figure~\ref{fig:conv2w} shows the convergence performance with 2 
workers. 

\begin{figure}[H]
        \centering
	\begin{subfigure}[b]{0.235\textwidth}
		\centering
		\includegraphics[width=\textwidth]{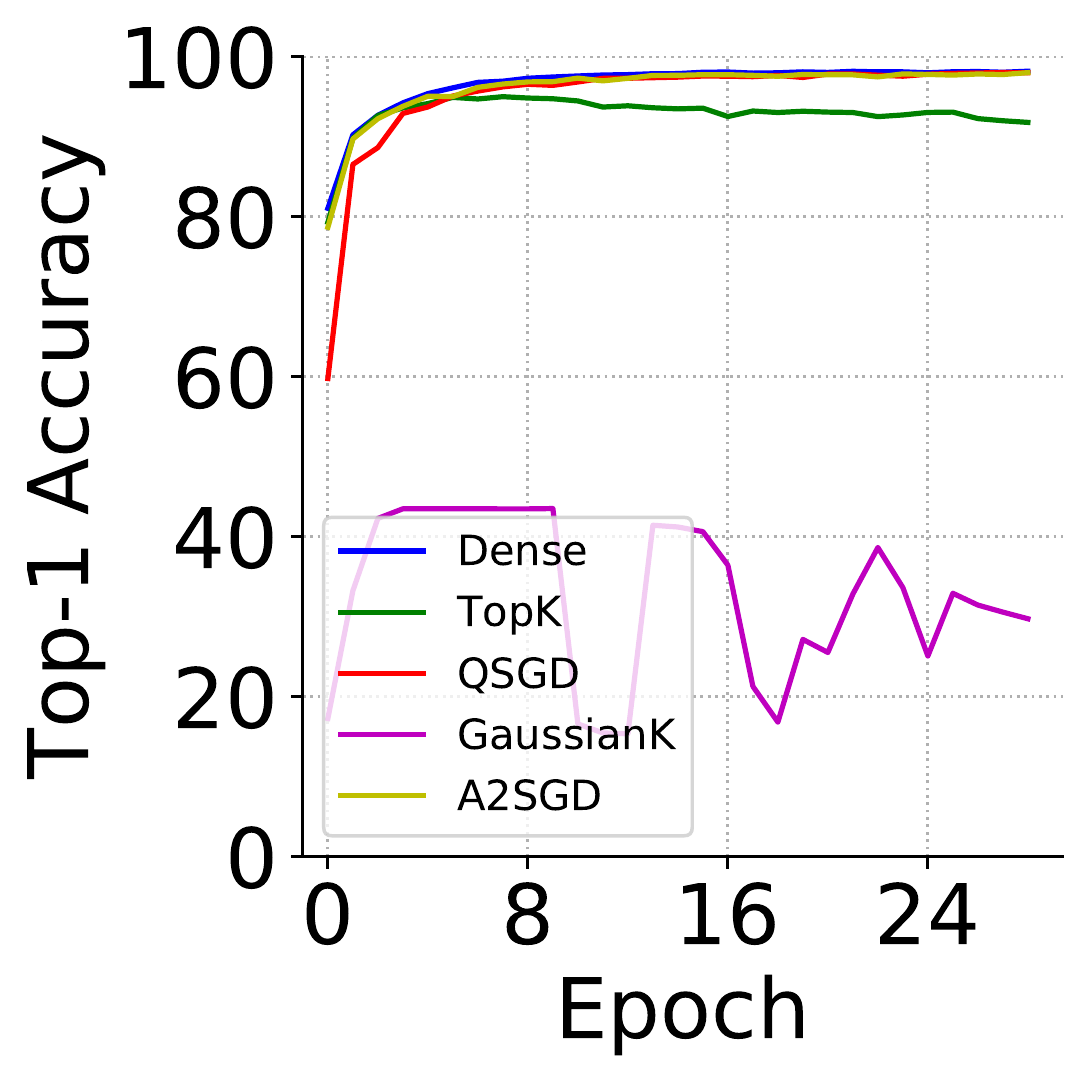}
		\caption{FNN3}
		\label{fig:FNN_acc}
	\end{subfigure}%
	~
	\begin{subfigure}[b]{0.215\textwidth}
		\centering
		\includegraphics[width=\textwidth]{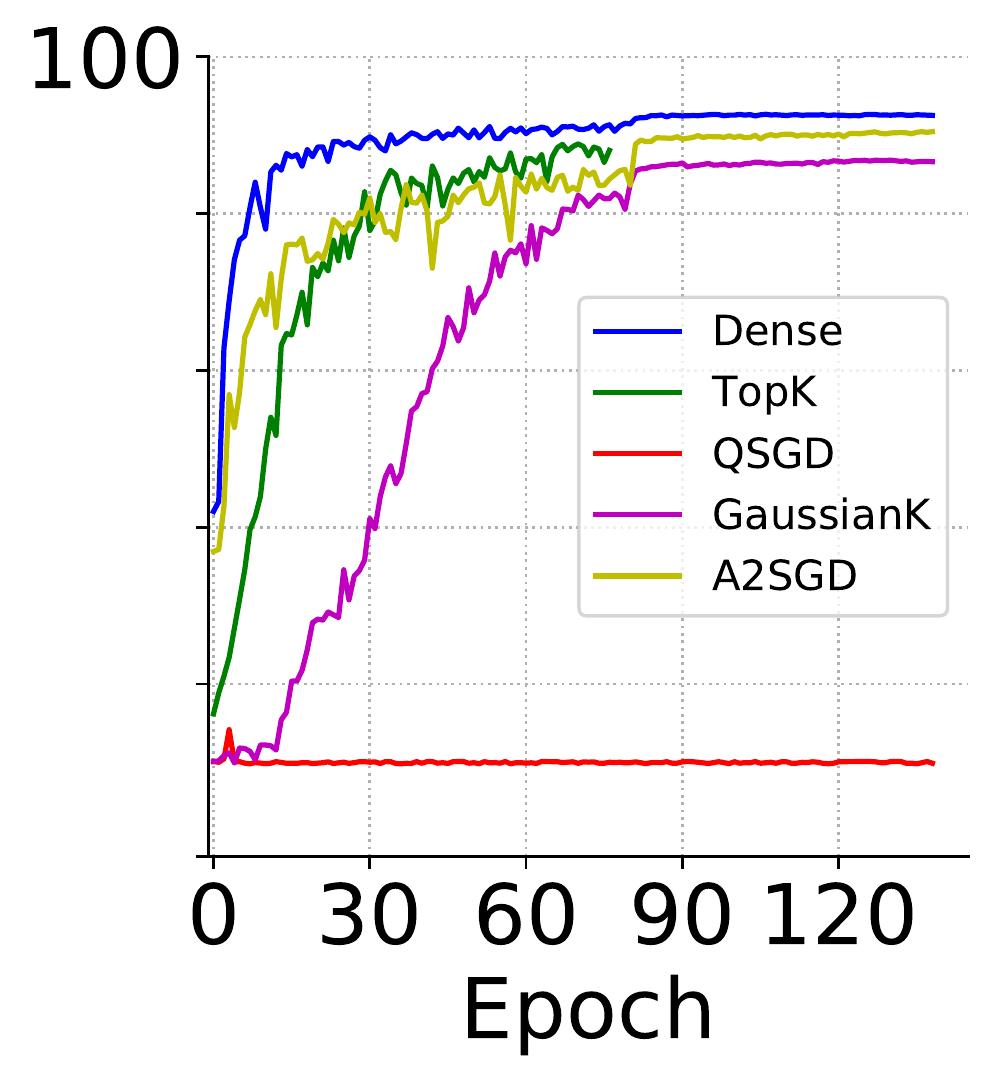}
		\caption{VGG16}
		\label{fig:VGG16_acc}
	\end{subfigure}%
	~
	\begin{subfigure}[b]{0.215\textwidth}
		\centering
		\includegraphics[width=\textwidth]{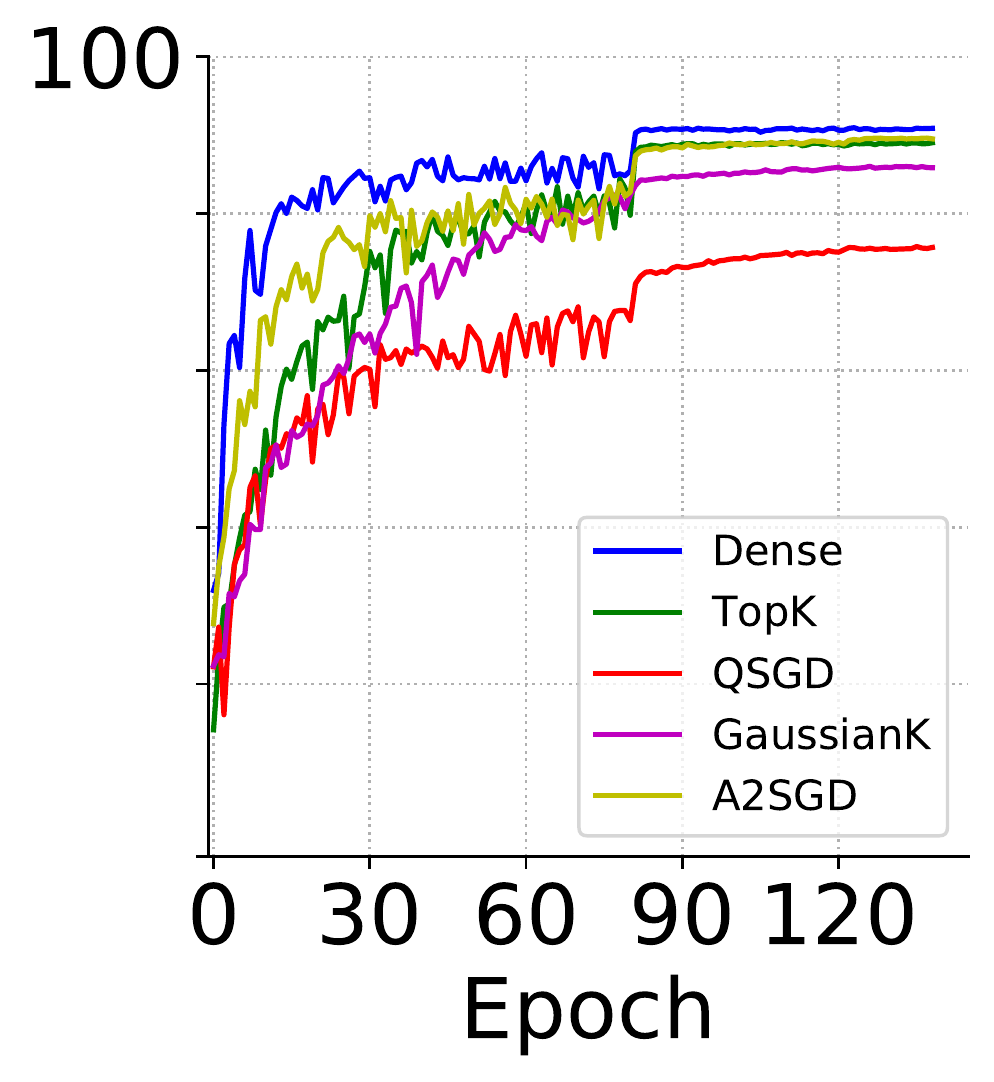}
		\caption{ResNet20}
		\label{fig:ResNet20_acc}
	\end{subfigure}%
	~
	\begin{subfigure}[b]{0.245\textwidth}
		\centering
		\includegraphics[width=\textwidth]{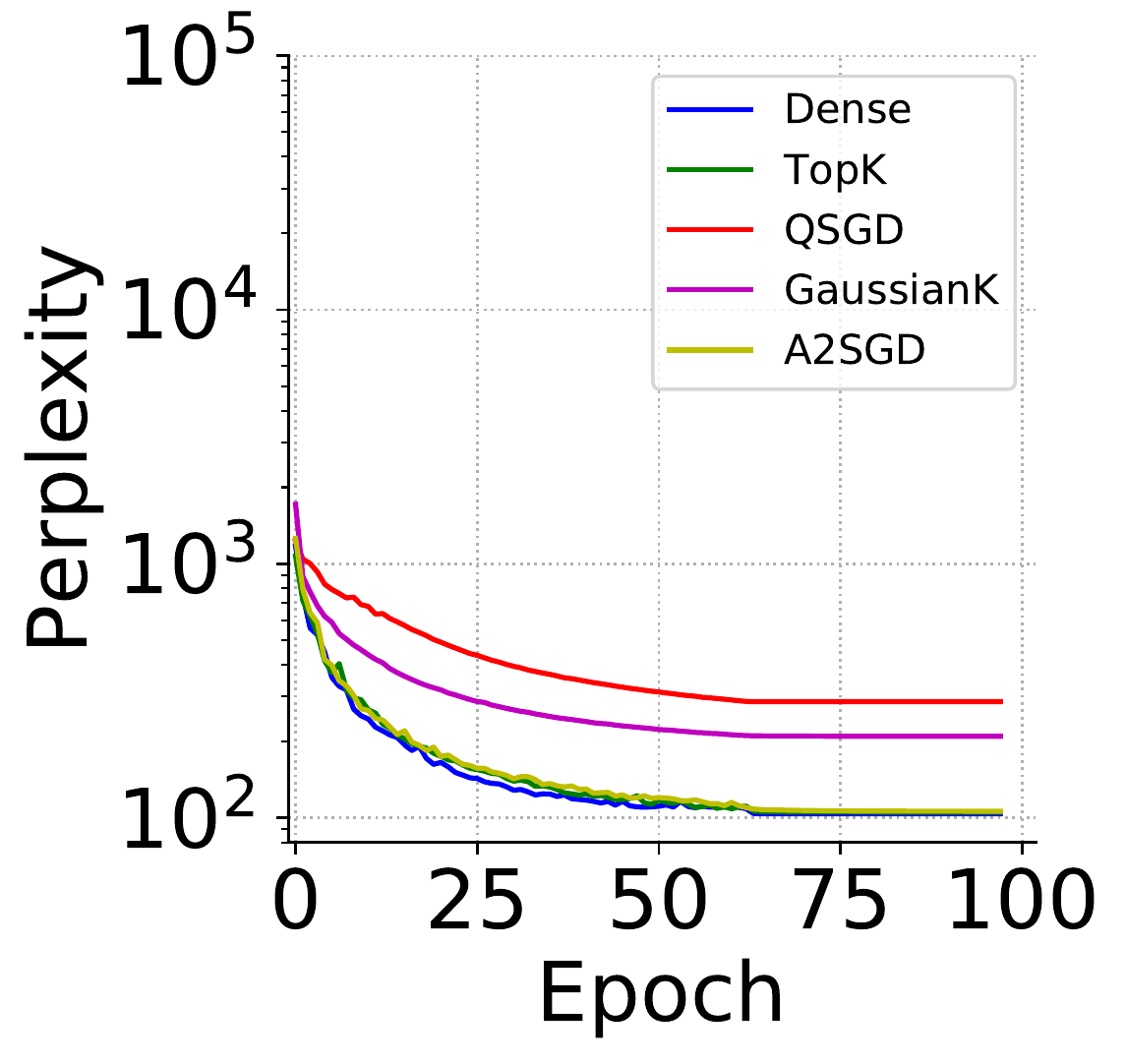}
		\caption{LSTM-PTB}
		\label{fig:LSTM_acc}
	\end{subfigure}%
	\caption{Comparison of Convergence Accuracy with 4 Workers}
	\label{fig:conv4w}
\end{figure}

Figure~\ref{fig:conv4w} shows the convergence performance with 4 workers. 

\begin{figure}[H]
	\centering
	\begin{subfigure}[b]{0.235\textwidth}
		\centering
		\includegraphics[width=\textwidth]{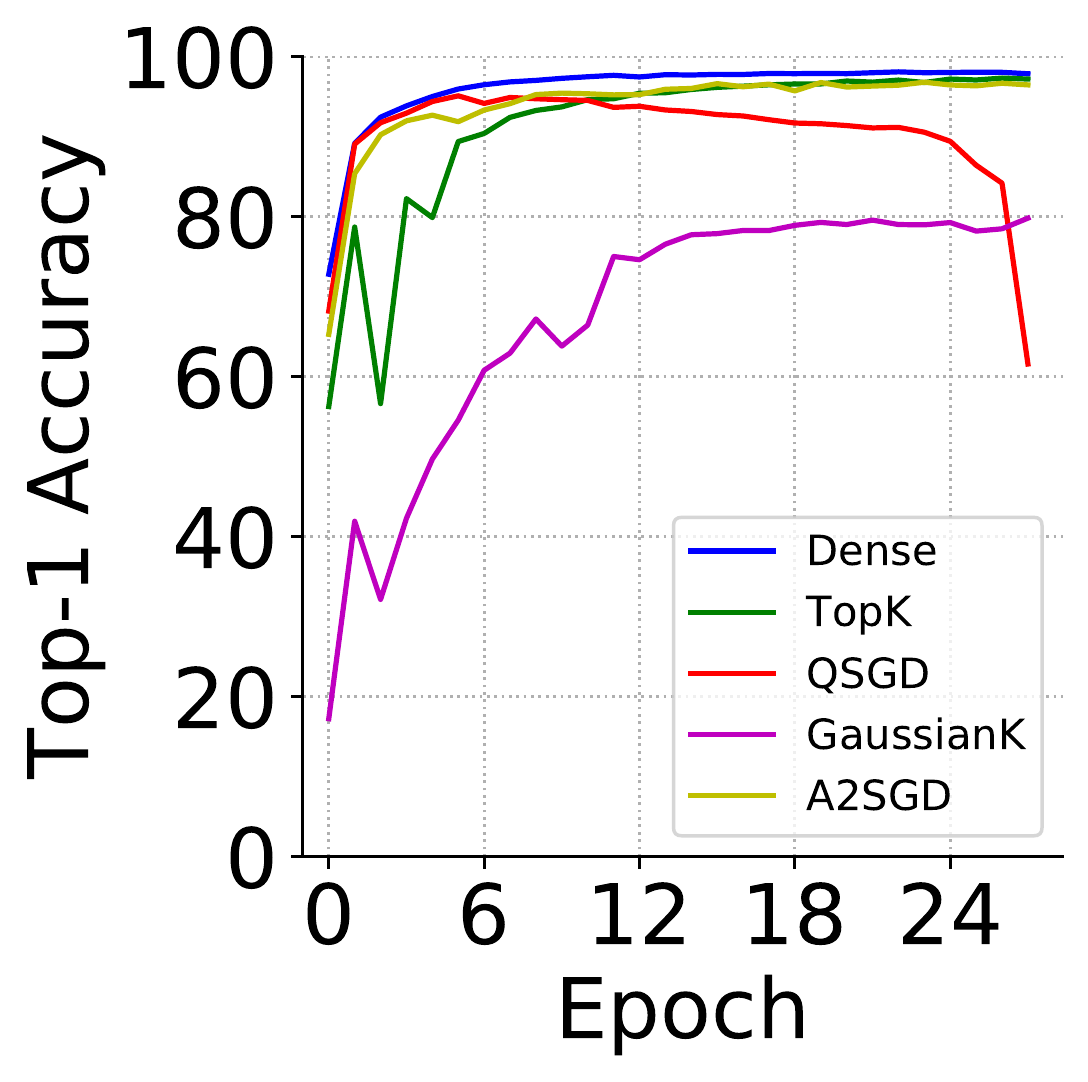}
		\caption{FNN3}
		\label{fig:FNN_acc}
	\end{subfigure}%
	~
	\begin{subfigure}[b]{0.215\textwidth}
		\centering
		\includegraphics[width=\textwidth]{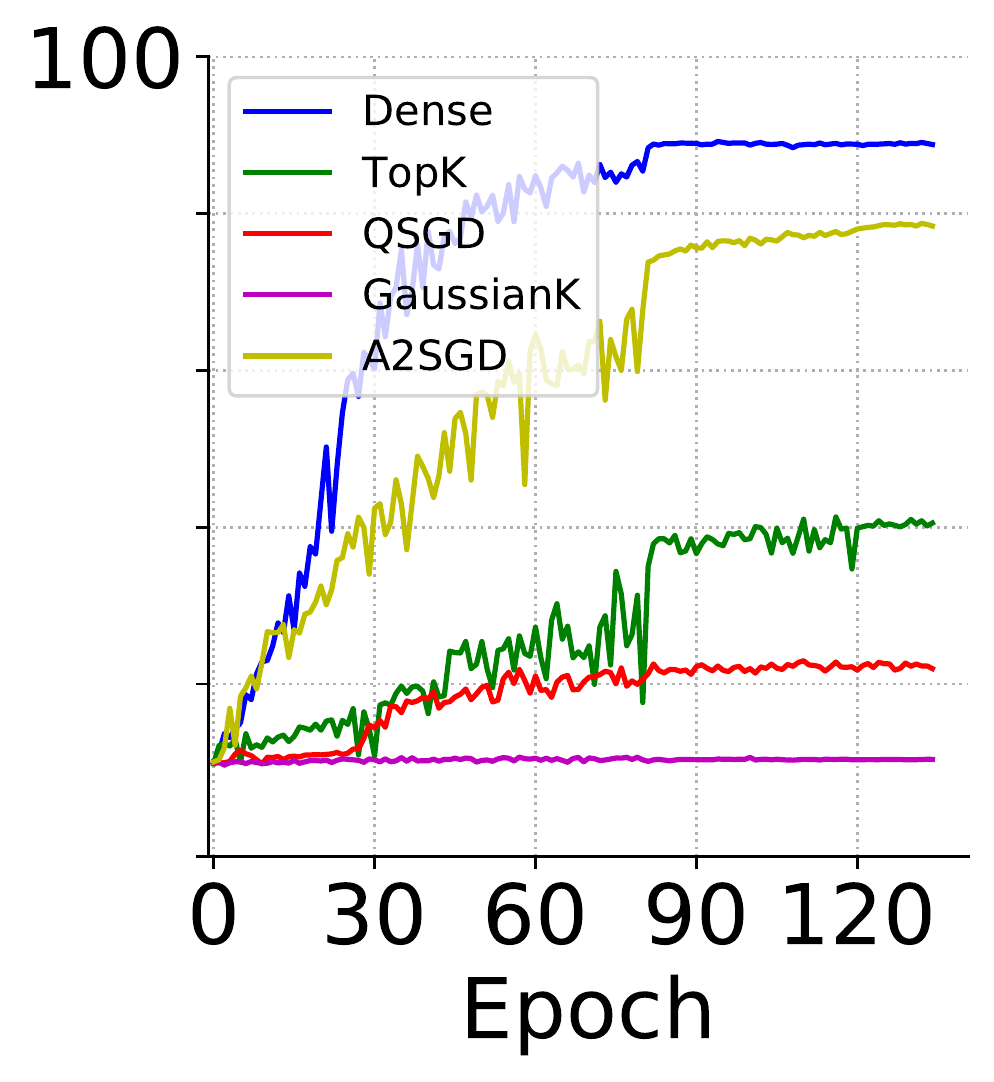}
		\caption{VGG16}
		\label{fig:VGG16_acc}
	\end{subfigure}%
	~
	\begin{subfigure}[b]{0.215\textwidth}
		\centering
		\includegraphics[width=\textwidth]{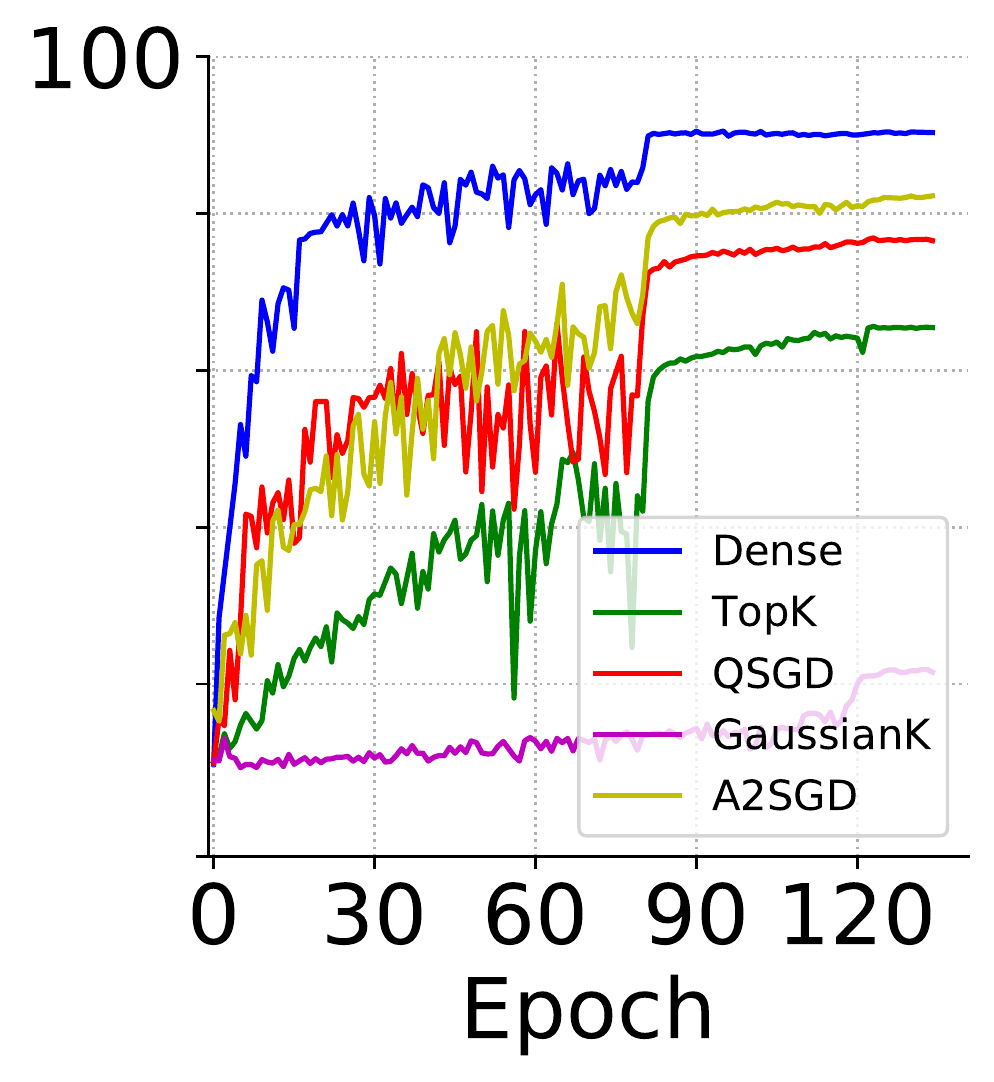}
		\caption{ResNet20}
		\label{fig:ResNet20_acc}
	\end{subfigure}%
	~
	\begin{subfigure}[b]{0.245\textwidth}
		\centering
		\includegraphics[width=\textwidth]{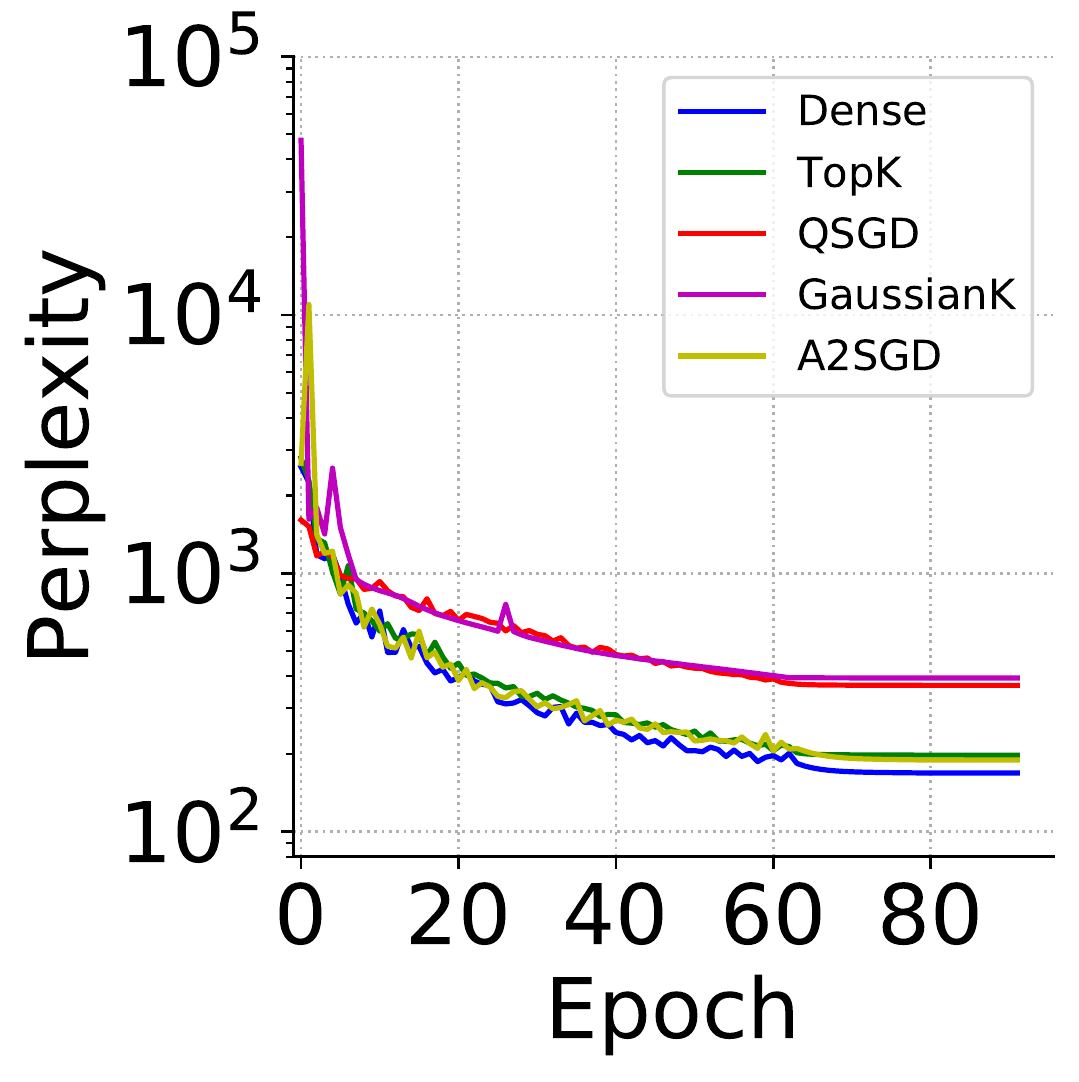}
		\caption{LSTM-PTB}
		\label{fig:LSTM_acc}
	\end{subfigure}
	\caption{Comparison of Convergence Accuracy with 16 Workers}
	\label{fig:conv16w}
\end{figure}

Figure~\ref{fig:conv16w} shows the convergence performance with 16 workers. 

\end{document}